\documentclass[11pt]{article}

\usepackage[final]{acl}

\usepackage{times}
\usepackage{latexsym}
\usepackage{graphicx}
\usepackage{subcaption}
\usepackage{booktabs}
\usepackage{multirow}
\usepackage{xcolor}
\usepackage{amsmath}
\usepackage[table]{xcolor}
\usepackage{adjustbox}
\usepackage{array}
\usepackage{CJKutf8}
\usepackage{tabularx}
\usepackage{enumitem}
\usepackage{ragged2e}
\usepackage{amssymb} 
\usepackage{amsfonts}
\usepackage{tcolorbox}
\usepackage{threeparttable}
\tcbuselibrary{breakable, skins}
\usepackage{graphicx}

\usepackage[T1]{fontenc}

\usepackage[utf8]{inputenc}

\usepackage{microtype}

\usepackage{inconsolata}

\usepackage{graphicx}
\title{RegioZH: Unveiling Comprehension Disparities across Regional Chinese Varieties in LLMs}

\author{
Weihua Zheng\textsuperscript{1,2}\quad
Wu Kui\textsuperscript{1}\quad
Thong T. Doan\textsuperscript{2} \\
\textbf{Lin Jingxia\textsuperscript{3}}\quad 
\textbf{AiTi Aw\textsuperscript{1}}\quad
\textbf{Nancy F. Chen\textsuperscript{1}}\quad
\textbf{Roy Ka-Wei Lee\textsuperscript{4}} \\
\textsuperscript{1}Institute of Advanced Intelligence and Computing (IAIC), A*STAR, Singapore \\
\textsuperscript{2}Singapore University of Technology and Design
\textsuperscript{3}Nanyang Technological University\\
\textsuperscript{4}University of British Columbia \\
\texttt{\{zheng\_weihua, wuk, aaiti, nancy\_chen\}@a-star.edu.sg} \\ \texttt{tienthong\_doan@sutd.edu.sg,\quad jingxialin@ntu.edu.sg} \\ \texttt{roy.lee@ubc.ca}
}

\begin{document}
\maketitle
\begin{abstract}

Current evaluations often treat Chinese varieties as monolithic, obscuring failures for regional users. We introduce RegioZH, a 2,607-item multiple-choice benchmark for regional lexical reasoning across Mainland China, Taiwan, Hong Kong, Singapore, and Malaysia. Built by region-specific annotators and expert-validated, it targets localized idioms, slang, code-mixing, politeness conventions, and sociocultural expressions. Zero-shot results on eleven LLMs show large disparities, including a 29.4-point average accuracy gap between Taiwan and Singapore. Persona and structured chain-of-thought prompts for five models yield selective gains, suggesting that some errors stem from retrieval or framing rather than only missing knowledge. Representation, attention, calibration, and surface-feature analyses show a dissociation between marker attention and semantic grounding, overconfident forced-choice errors, and localized lexical density as a consistent difficulty predictor. Qualitative analysis finds terminology misunderstandings, sociocultural inference gaps, and cross-regional semantic divergence. RegioZH demonstrates that apparent Chinese proficiency can mask uneven understanding across regional varieties, motivating intra-language robustness evaluation alongside multilingual generalization.

\textbf{Dataset}: \url{https://github.com/Social-AI-Studio/RegioZH}

\end{abstract}

\section{Introduction}

\begin{figure*}[t!]
  \centering
  \captionsetup{skip=0pt}
  \includegraphics[width=\textwidth]{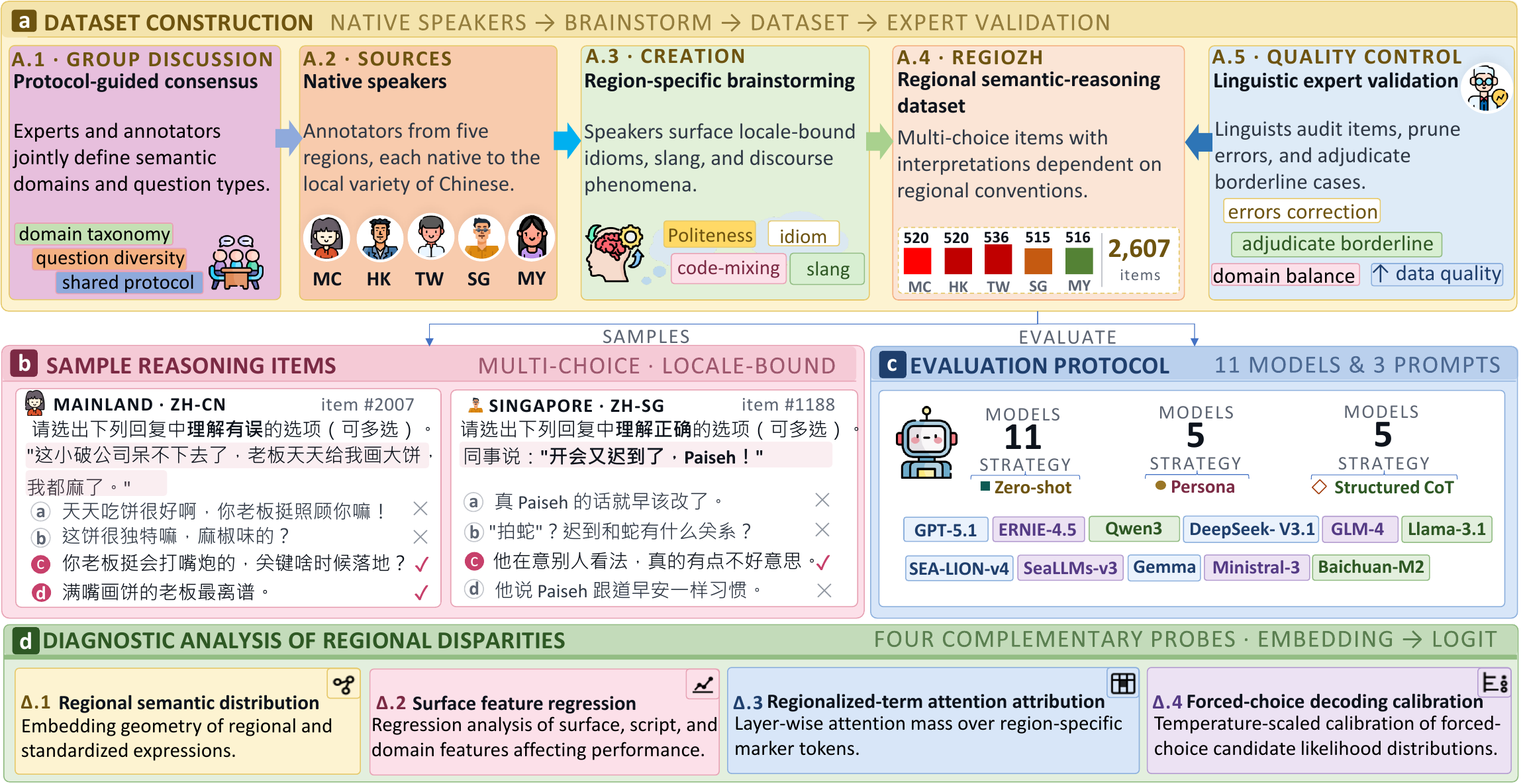}
  \caption{Evaluation framework overview: data creation pipeline, dataset samples, and evaluation dimensions.}
  \label{fig:pipeline}
\end{figure*}

Large language models are routinely described as supporting Chinese, yet this framing obscures a fundamental heterogeneity. Chinese is instantiated across regional varieties whose lexical inventories, institutional conventions, and contact-induced written features diverge in ways consequential for real users~\cite{grieve2025sociolinguistic}. Speakers across Chinese-speaking regions routinely employ region-specific expressions that are natural within their local communities but opaque to outsiders. Robust Chinese understanding therefore demands more than proficiency in standardized Mandarin; it requires reasoning over region-sensitive lexical and sociocultural cues.

This challenge exemplifies a broader problem in intra-language robustness~\citep{hofmann2024dialect, sap2019risk, grandury2025leaderboard}. Prior work has documented isolated disparities in Chinese~\citep{ lyu2025characterizing, chen2024measuring, jiang2025well, cheng2025hkcanto}, yet leaves unclear whether models can use localized lexical cues for region-appropriate inference. 
Chinese is a uniquely informative testbed because script differences, regional lexical and institutional conventions, and code-switching patterns~\citep{huzaifah2024evaluating} coexist within one linguistic space, enabling the simultaneous diagnosis of failure mechanisms that aggregate evaluations typically mask.

To address this gap, we introduce \textsc{RegioZH} (Fig.~\ref{fig:pipeline}), a 2{,}607-item benchmark for \textit{regional lexical reasoning} in Chinese covering Mainland China, Taiwan, Hong Kong, Singapore, and Malaysia. Each region-labeled multiple-choice item asks models to select the interpretation licensed by localized cues, including idioms, politeness conventions, and sociocultural expressions. This format tests whether models map localized forms to region-appropriate meanings rather than standard Chinese proficiency. Items are authored by region-specific annotators and validated by Chinese linguistics experts. In zero-shot evaluation of 11 representative LLMs, we observe a near 30-point accuracy disparity between the highest- and lowest-performing regions, indicating that nominal Chinese capability can mask substantial internal unevenness. 

The benchmark also enables us to probe a second technical gap: aggregate errors do not reveal whether regional failures reflect missing knowledge, poor elicitation, or miscalibrated interpretation. Through persona and structured chain-of-thought prompting~\citep{wei2022chain}, we find that some region-specific knowledge can be latent but unreliably activated under standard settings. This corroborates recent findings that reasoning trajectories significantly modulate multilingual factual reasoning performance~\citep{weihua2026adamcot}. We further apply representation geometry, attention attribution, forced-choice calibration, and surface-feature regression as complementary diagnostic lenses. These analyses suggest that evaluated LLMs often attend to localized markers without grounding them in region-appropriate meanings, while exhibiting high confidence in erroneous regional interpretations.

We summarize our contributions as follows:

\begin{itemize}[leftmargin=*,topsep=0pt]
\setlength{\itemsep}{0pt}
\setlength{\parskip}{0pt}
\setlength{\parsep}{0pt}
    \item \textbf{Native-authored regional benchmark.} We introduce \textsc{RegioZH}, a five-region dataset designed to diagnose regional lexical reasoning in LLMs.
    \item \textbf{Knowledge elicitation diagnosis.} Prompt gains suggest that regional knowledge is partly latent, with performance gaps reflecting elicitation failures rather than absolute deficits.
    \item \textbf{Four-lens diagnostic analysis.} We provide diagnostic evidence of grounding and calibration breakdowns, with localized lexical count as a salient difficulty predictor.
    \item \textbf{Fine-grained error typology.} We identify terminology misunderstanding, semantic divergence, and sociocultural inference as core bottlenecks.
\end{itemize}
\noindent

\section{Related Work}

\textbf{Chinese and Sinophone LLM evaluation.} Existing Chinese evaluation benchmarks have made substantial contributions to assessing general-purpose language understanding and disciplinary reasoning. General-purpose Chinese evaluation benchmarks~\citep{huang2023c, li2024cmmlu, zhong2024agieval, zhang2023m3exam} treat Chinese as a linguistically unified target, providing limited coverage of regional lexical and sociocultural variation. Recent efforts have begun addressing specific Chinese varieties: TMLU~\citep{chen2024measuring} and TMMLU+~\citep{tam2024tmmlu+} reveal systematic disparities in Taiwanese Mandarin understanding, \citet{lyu2025characterizing} demonstrated task-dependent biases between Simplified and Traditional Chinese regional terms, and HKCanto-Eval~\citep{cheng2025hkcanto} and Yue-Benchmark~\citep{jiang2025well} expose persistent gaps in Cantonese linguistic and cultural comprehension. These studies establish that "Chinese-capable" LLMs do not perform uniformly across varieties, yet each focuses on a single region or binary script contrast. Our work addresses this gap by constructing a unified benchmark spanning five regional varieties and diagnosing failure mechanisms that operate across them simultaneously.

\textbf{Intra-language robustness.} Multilingual alignment work targets \textit{inter}-language gaps through cross-lingual mapping objectives and alignment metrics~\citep{zheng2026bridging,zheng2025ccl}. Intra-language variation is nominally one language, so its alignment failures remain invisible in aggregate multilingual evaluation. Studies on AAVE~\citep{blodgett2016demographic,sap2019risk} and ReDial~\citep{lin2025assessing} demonstrate that dialectal surface variation induces performance drops exceeding 10\% across GPT, Claude, and Llama model families, even under chain-of-thought prompting, suggesting a gap between knowledge storage and reliable knowledge retrieval~\citep{yu2026compositioncollapsestablefactual}. Similar findings hold for Spanish regional varieties~\citep{grandury2025leaderboard}. Unlike ReDial and related work, which test robustness to dialectal rewrites of existing tasks, our benchmark is built around naturally occurring region-specific lexical items curated by native annotators, targeting regional lexical reasoning rather than semantic-preserving surface robustness. Additional related work is discussed in ~\ref{sec:related work}.


\section{Dataset Construction}
\label{sec:dataset creation}

As shown in Figure~\ref{fig:pipeline}, RegioZH was built through a protocol-guided design--creation--review pipeline. The task is multi-select multiple-choice regional lexical/semantic interpretation in written Chinese: given a conversational scenario with a region-specific expression, the model judges candidate responses based on contextual regional understanding. Focus groups with linguistics experts and regional annotators defined the semantic domains and question types. Each item contains a Chinese stem, a target region, answer options with one or more gold answers, domain/question-type annotations, and a validation rationale. Accuracy uses exact set match over selected options; explanations, when elicited, are analyzed separately.

Regional annotators created items using the shared taxonomy and checked local sources for region-specific meanings, usage contexts, and sociocultural assumptions. Items cover locale-bound idioms, code-mixing, politeness norms, institutional references, and culturally specific expressions across Mainland China, Taiwan, Hong Kong, Singapore, and Malaysia. We initially targeted $\sim$550 items per region. Three Chinese-linguistics experts independently reviewed each candidate item for target-phenomenon fit and answer-key unambiguity. For Fleiss' $\kappa$, the rated unit was the candidate item and the label space was the pre-adjudication validation decision: retain, revise, or remove, yielding substantial agreement (Fleiss' $\kappa = 0.762$). Disagreements were adjudicated through expert discussion, and only validated or revised-and-validated items were retained.

The final dataset contains 2{,}607 validated multiple-choice items. This process is intended to mitigate single-contributor bias and improve coverage, while RegioZH remains a diagnostic sample rather than an exhaustive account of within-region variation or overall Chinese proficiency. Further details are provided in~\ref{sec:dataset creation}.

\section{Experiment Setup}

We evaluate 11 LLMs on \textit{RegioZH}, grouping them by publicly stated regional or language focus rather than unverifiable corpus coverage: Chinese-centric models (ERNIE-4.5-21B, Qwen3-30B, DeepSeek-V3.1, GLM-4-32B, Baichuan-M2-32B), Southeast Asia--oriented models (SEA-LION-v4-27B, SeaLLMs-v3-7B), and global general-purpose models developed outside the Chinese ecosystem (GPT-5.1, Llama-3.1-8B, Gemma-7B, Ministral-3-14B). This descriptive grouping supports comparison across stated regional priors for localized Chinese lexical interpretation.

We report zero-shot accuracy for all models. Persona and structured chain-of-thought (CoT) prompts are additionally run on five models spanning the zero-shot performance range (DeepSeek-V3.1, GPT-5.1, Qwen3-30B, Llama-3.1-8B, Ministral-3-14B), testing whether performance changes when an expert role is specified or regional cues are made explicit, without treating such changes as evidence of training-data coverage. Full implementation details are in Appendix~\ref{sec:Experiment setting and Prompt}; model usage is summarized in ~\ref{sec:model_sum}.



\section{Experimental Results}
\subsection{Zero-Shot Disparities across Regional Chinese Varieties}
\begin{table*}[t!]
\centering
\footnotesize
\setlength{\tabcolsep}{2.3pt}

\begin{tabular}{l|ccccc}
\hline
\textbf{Model} & \textbf{Malaysia} & \textbf{Hong Kong} & \textbf{Mainland} & \textbf{Singapore} & \textbf{Taiwan} \\
\hline
DeepSeek-V3.1 & \textbf{73.6} $_{2.2}$ $[69.1, 77.6]$ & \textbf{80.8} $_{1.7}$ $[77.2, 83.9]$ & 78.8 $_{1.8}$ $[75.1, 82.1]$ & 42.7 $_{2.2}$ $[38.5, 47.0]$ & \textbf{81.9} $_{1.7}$ $[78.4, 84.9]$ \\
GPT-5.1 & 71.9 $_{2.2}$ $[67.4, 76.0]$ & 61.5 $_{2.1}$ $[57.3, 65.6]$ & \textbf{81.0} $_{1.7}$ $[77.4, 84.1]$ & \textbf{57.7} $_{2.2}$ $[53.4, 61.9]$ & 77.6 $_{1.8}$ $[73.9, 80.9]$ \\
Qwen3-30B & 60.8 $_{2.4}$ $[56.0, 65.4]$ & 66.2 $_{2.1}$ $[62.0, 70.1]$ & 69.2 $_{2.0}$ $[65.1, 73.0]$ & 36.5 $_{2.1}$ $[32.5, 40.8]$ & 75.6 $_{1.9}$ $[71.8, 79.0]$ \\
Baichuan-M2 & 63.2 $_{2.4}$ $[58.5, 67.7]$ & 47.9 $_{2.2}$ $[43.6, 52.2]$ & 74.6 $_{1.9}$ $[70.7, 78.2]$ & 10.7 $_{1.4}$ $[8.3, 13.6]$ & 31.2 $_{2.0}$ $[27.4, 35.2]$ \\
Ernie-4.5 & 51.9 $_{2.5}$ $[47.1, 56.7]$ & 52.9 $_{2.2}$ $[48.6, 57.1]$ & 58.3 $_{2.2}$ $[54.0, 62.4]$ & 28.4 $_{2.0}$ $[24.6, 32.4]$ & 67.7 $_{2.0}$ $[63.6, 71.5]$ \\
Gemma-7B & 12.5 $_{1.6}$ $[9.7, 16.0]$ & 6.3 $_{1.1}$ $[4.5, 8.8]$ & 7.1 $_{1.1}$ $[5.2, 9.7]$ & 17.1 $_{1.7}$ $[14.1, 20.6]$ & 20.5 $_{1.7}$ $[17.3, 24.1]$ \\
SEA-LION-27B & 61.5 $_{2.4}$ $[56.8, 66.1]$ & 8.5 $_{1.2}$ $[6.4, 11.2]$ & 67.7 $_{2.0}$ $[63.6, 71.6]$ & 36.9 $_{2.1}$ $[32.8, 41.1]$ & 74.2 $_{1.9}$ $[70.4, 77.8]$ \\
GLM-4-32B & 68.3 $_{2.3}$ $[63.7, 72.8]$ & 39.8 $_{2.1}$ $[35.7, 44.1]$ & 66.0 $_{2.1}$ $[61.8, 69.9]$ & 39.8 $_{2.2}$ $[35.7, 44.1]$ & 71.8 $_{1.9}$ $[67.9, 75.5]$ \\
Ministral-3-14B & 54.3 $_{2.4}$ $[49.5, 59.0]$ & 35.6 $_{2.1}$ $[31.6, 39.8]$ & 54.6 $_{2.2}$ $[50.3, 58.9]$ & 32.4 $_{2.1}$ $[28.5, 36.6]$ & 67.3 $_{2.0}$ $[63.3, 71.2]$ \\
SeaLLMs-v3-7B & 49.3 $_{2.4}$ $[44.5, 54.1]$ & 10.2 $_{1.3}$ $[7.6, 12.8]$ & 57.7 $_{2.2}$ $[53.4, 62.0]$ & 30.3 $_{2.0}$ $[26.4, 34.2]$ & 61.0 $_{2.1}$ $[56.9, 65.1]$ \\
Llama-3.1-8B & 28.4 $_{2.2}$ $[24.2, 32.9]$ & 23.8 $_{1.9}$ $[20.4, 27.7]$ & 23.8 $_{1.9}$ $[20.4, 27.7]$ & 14.4 $_{1.5}$ $[11.6, 17.7]$ & 41.6 $_{2.1}$ $[37.5, 45.8]$ \\
\hline
\textbf{Average} & 54.2 $_{2.5}$ $[49.3, 59.0]$ & 39.4 $_{2.1}$ $[35.3, 43.5]$ & 58.1 $_{2.2}$ $[53.8, 62.2]$ & \cellcolor{green!20} 31.5 $_{2.0}$ $[27.6, 35.5]$ & \cellcolor{blue!20} 60.9 $_{2.1}$ $[56.9, 65.0]$ \\
\hline
\textbf{Human} & 78.1 $_{1.8}$ $[74.6, 81.7]$ & 81.5 $_{1.7}$ $[78.2, 84.9]$ & 85.6 $_{1.5}$ $[82.6, 88.6]$ & 79.5 $_{1.8}$ $[76.0, 83.0]$ & 83.4 $_{1.6}$ $[80.3, 86.6]$ \\
\hline
\end{tabular}
\caption{Model zero-shot performance on \textit{RegioZH}, measured by Accuracy (\%). ``-'' means ``not support''. Results are presented as Mean $_{SE}$ $[95\% \text{ CI}]$, where the bolded figure is the primary metric, followed by its standard error in subscript and the confidence interval in brackets. The best-performing model in each region is \textbf{bolded}.}
\label{tab:zero-shot-result}
\end{table*}

Tab.~\ref{tab:zero-shot-result} shows substantial cross-regional disparities on \textit{RegioZH}, indicating that region-licensed interpretation of localized Chinese expressions is not captured by a single aggregate Chinese score. Averaged across models, Taiwan leads at 60.9\%, followed by Mainland China (58.1\%) and Malaysia (54.2\%), while Hong Kong (39.4\%) and Singapore (31.5\%) lag behind. The 29.4-point Taiwan--Singapore gap shows that region-sensitive interpretation remains uneven under zero-shot prompting.

This unevenness appears across model ecosystems. Chinese-centric models perform strongest overall, with DeepSeek topping three of five regions and Qwen3 remaining comparatively robust, yet both degrade on Singapore and Hong Kong. GPT-5.1 leads among global models on Mainland China and Singapore, whereas Southeast Asia-oriented models show no clear advantage on these localized items. At the model level, several systems also show large row-wise swings; for example, SEA-LION-27B reaches 74.2\% on Taiwan but drops to 8.5\% on Hong Kong. Overall, zero-shot results suggest unreliable mappings from localized surface markers to region-appropriate interpretations, without distinguishing knowledge gaps from retrieval, prompting, or calibration effects. Additional analyses controlling for surface artifacts and cross-regional differences in item structure, including option count and gold-set cardinality, indicate that neither factor fully explains the observed regional disparities; see Appendix~\ref{sec:Supplementary Quantitative Results} for details.


\subsection{Eliciting Regional Knowledge with Persona and Structured Reasoning}
\label{Eliciting Regional Knowledge with Persona and Structured Reasoning}
\begin{figure*}[t!]
\centering
\captionsetup{skip=0pt}
\includegraphics[width=0.98\textwidth]{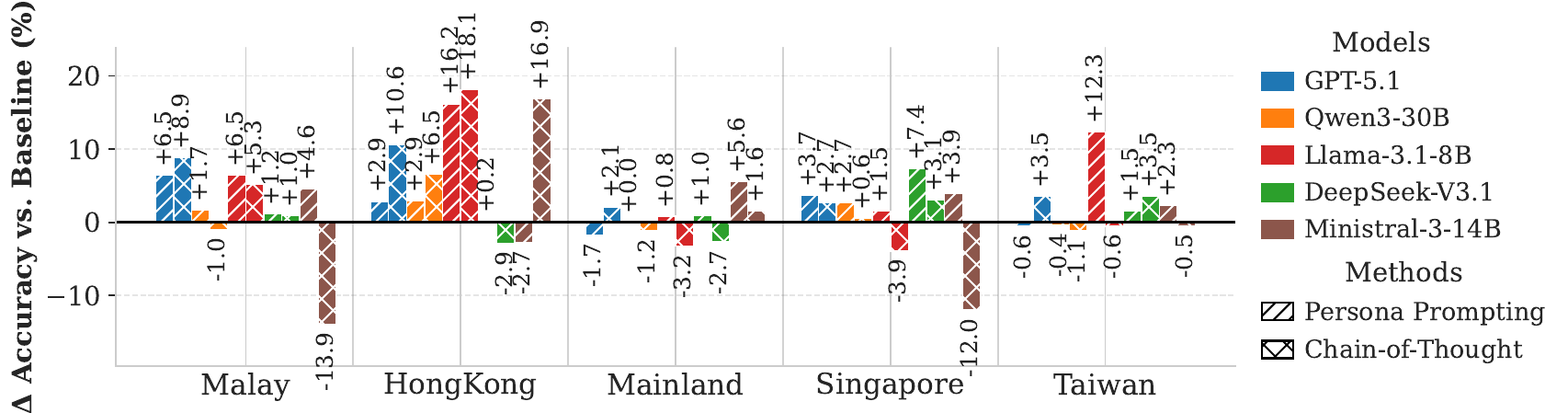}
\caption{Regional accuracy gains from persona and structured CoT prompting across five models on \textit{RegioZH}.}
\label{fig:comparison}
\end{figure*}
To examine prompt sensitivity, we apply persona and structured CoT prompting to five models under matched decoding and parsing settings. Only the final multiple-choice option is scored, so gains are diagnostic and may reflect changes in answer priors, parsing success, verbosity, regional salience, or elicitation of relevant knowledge.

Fig.~\ref{fig:comparison} shows that persona prompting is more stable, improving accuracy in 20 of 25 model--region combinations versus 14 for structured CoT, with positive average gains for all five models. Average persona gains are largest for Hong Kong (+3.90), Malaysia (+3.88), and Singapore (+3.84), followed by Taiwan (+2.80) and Mainland China (+1.14), broadly aligning with lower zero-shot performance in regions requiring stronger localized lexical grounding. 

Structured CoT is more polarized: it improves GPT-5.1 across all regions, with large margins on Hong Kong and Malaysia, but yields mixed results for other models, including sharp drops for Ministral-3-14B on Malaysia and Singapore. Overall, regional linguistic reasoning failures are partly prompt-sensitive: expert persona prompts may activate region-specific knowledge, while structured CoT depends more on accurate early grounding and the model’s ability to integrate localized cues into subsequent reasoning.

\section{Diagnostic Analyses of Regional Failures}


Prompting results suggest that regional failures are sensitive to elicitation strategy, reflecting unstable knowledge elicitation and localized lexical-reasoning difficulties rather than absolute knowledge deficits alone. We examine four correlational diagnostic signals: representation spaces, attention allocation, decoding calibration, and inter-domain interference, to test whether RegioZH errors co-vary with failures to encode, attend to, calibrate, or compose region-licensed interpretations. The focus is not general Chinese proficiency, but whether models can connect localized expressions, regional context, and licensed meanings at answer time.

We conduct these analyses mainly zero-shot, using Qwen3-30B, GLM-4-32B, and SeaLLMs-v3-7B as top-, mid-, and lower-tier open-source models. Because patterns are consistent across them, Sections~\ref{sec:Representation Geometry}, ~\ref{sec:Attention without Grounding}, and ~\ref{sec:Confidence without Understanding} mainly report GLM results, with full Qwen and SeaLLM results in~\ref{sec: Results of Qwen and SeaLLMs}.


\subsection{Representation Geometry and Regional Separation}
\label{sec:Representation Geometry}



To examine whether regional-understanding failures coincide with representational shifts, we project final-layer sentence embeddings into a shared 3D PCA space and quantify pairwise distributional discrepancies using the unbiased empirical estimate of squared Maximum Mean Discrepancy (MMD$^2$), a kernel-based measure of the difference between two distributions. Within the same representation and kernel configuration, smaller MMD$^2$ values indicate greater relative similarity between embedding distributions, whereas larger values indicate greater distributional separation. Such shifts may reflect both orthographic differences, such as those between Simplified and Traditional Chinese, and genuine lexical, syntactic, or regional-expression differences. To partially disentangle these factors, we construct human-verified Simplified and Traditional formalized versions for Hong Kong and Taiwan, whose original informal data are exclusively in Traditional Chinese (see details in~\ref{sec: Results of Qwen and SeaLLMs}). This analysis separates representational proximity from downstream correctness: close embedding distributions do not guarantee correct answers, whereas large distributional shifts may reveal localized forms that are encoded sufficiently differently to become diagnostically informative.

\begin{figure}[t!]
  \centering
  \captionsetup{skip=0pt}
  \includegraphics[width=0.49\textwidth]{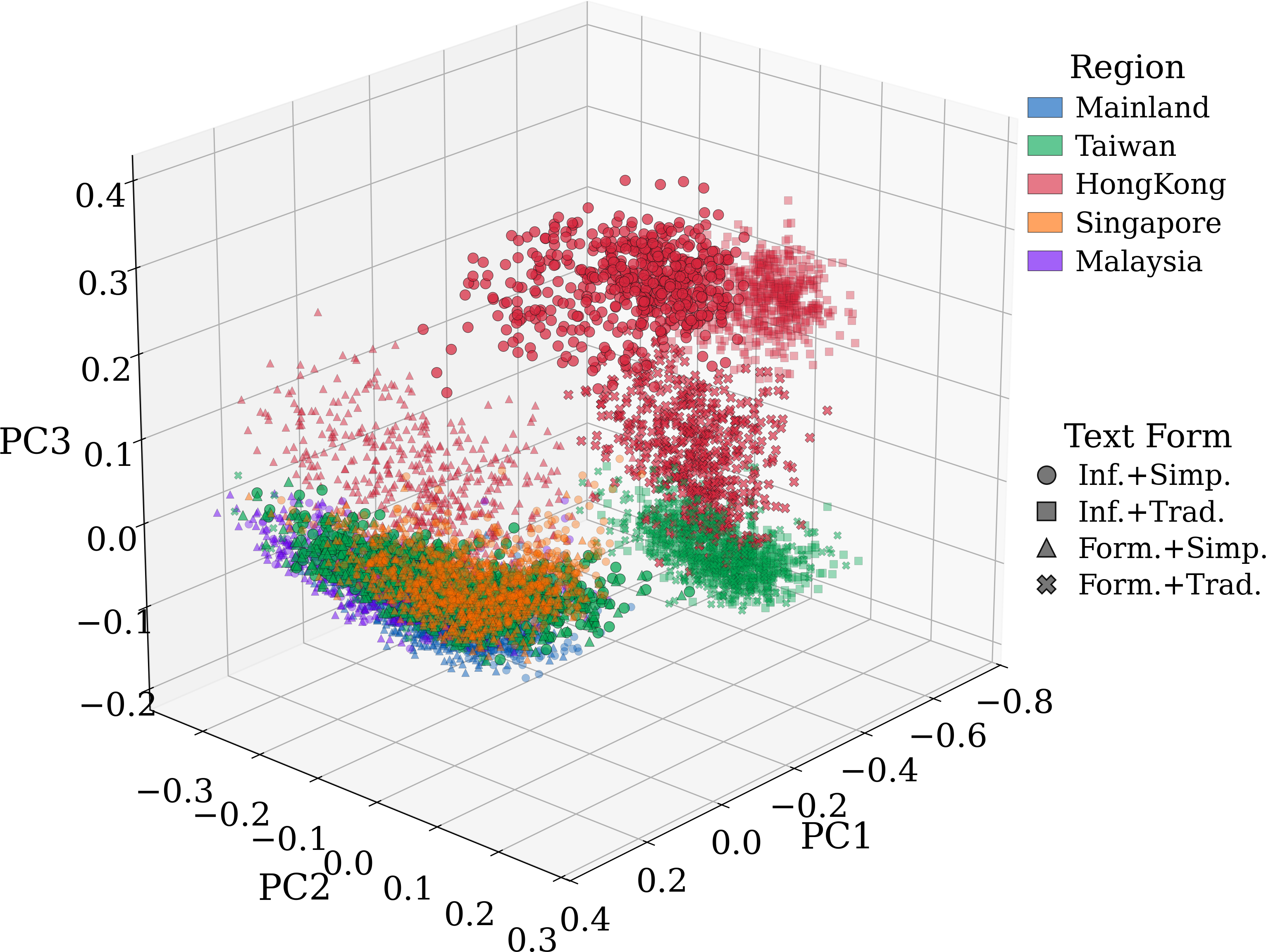}
  \caption{PCA projection of regional expressions and their standardized counterparts.Inf., Form., Trad., and Simp. denote informal region-specific expressions, formal standard expressions, traditional script, and simplified script, respectively.}
  \label{fig:semantic_representation}
\end{figure}

Fig.~\ref{fig:semantic_representation} shows that script structures the model's semantic space. Except for Hong Kong informal expressions, MMD$^2$ values(~\ref{tab:mmd2}) between each region's formal/informal data and other regions' Simplified data remain below $0.15$, indicating proximity. Converting Hong Kong informal expressions to Simplified reduces their gap with other regions' informal Simplified data from $1.215$ to $0.75$, while Taiwan also moves closer to other regions under Simplified conversion. Thus, script is salient but not the only dimension affecting regional comprehension.

Lexical and syntactic factors are important for Hong Kong, whose informal expressions reflect Cantonese vocabulary and syntax in addition to Traditional script. When Hong Kong and Taiwan informal data are formalized and uniformly converted to Simplified, Hong Kong's MMD$^2$ distance to other regions drops from $0.75$ to $0.11$, with similar convergence elsewhere. This suggests that benchmark difficulty partly lies in mapping informal local forms to region-licensed meanings, rather than processing Chinese text in general.

Overall, regional Chinese data are closest in semantic space under formal expression and a unified writing system, but Taiwanese and Hong Kong Traditional Chinese should not be treated as the same surface form. Taiwan remains robust: it achieves the highest average performance under Traditional input, and Sec.~\ref{sec: Surface Feature Interference} shows Simplified--Traditional conversion has only marginal downstream impact. Hong Kong performance, by contrast, appears more affected by Cantonese syntax, Hong Kong-specific Traditional conventions, and regional lexical items, which may be less represented in Chinese web corpora. Thus, representational separation is most useful when interpreted with task accuracy, helping distinguish orthographic sensitivity, localized lexical mapping, and variety-specific structural divergence without itself determining correctness.

\begin{table}[htpb]
\centering
\small
\setlength{\tabcolsep}{3pt}
\renewcommand{\arraystretch}{0.85}
\begin{tabular}{l ccccc cc}
\toprule
& \multicolumn{5}{c}{Simplified} & \multicolumn{2}{c}{Traditional} \\
\cmidrule(lr){2-6} \cmidrule(lr){7-8}
Register & HK & TW & SG & MC & MY & HK & TW \\
\midrule
Informal & 0.750 & 0.090 & 0.115 & 0.145 & 0.119 & 1.215 & 0.724 \\
Formal   & 0.110 & 0.050 & 0.070 & 0.080 & 0.070 & 1.085 & 1.140 \\
\bottomrule
\end{tabular}
\caption{Cross-regional MMD$^2$ distances. Each cell reports the unbiased MMD$^2$ between the focal region (column) and the pooled distribution of the remaining four regions, under the matching register. The reference distribution is always in Simplified; the right-hand panel keeps HK and TW in Traditional. Lower values indicate greater proximity to the rest of the corpus.}
\label{tab:mmd2}
\end{table}


\subsection{Attention–Performance Mismatch}
\label{sec:Attention without Grounding}

\begin{figure}[t]
  \centering
  \captionsetup{skip=0pt}
  \includegraphics[width=0.49\textwidth]{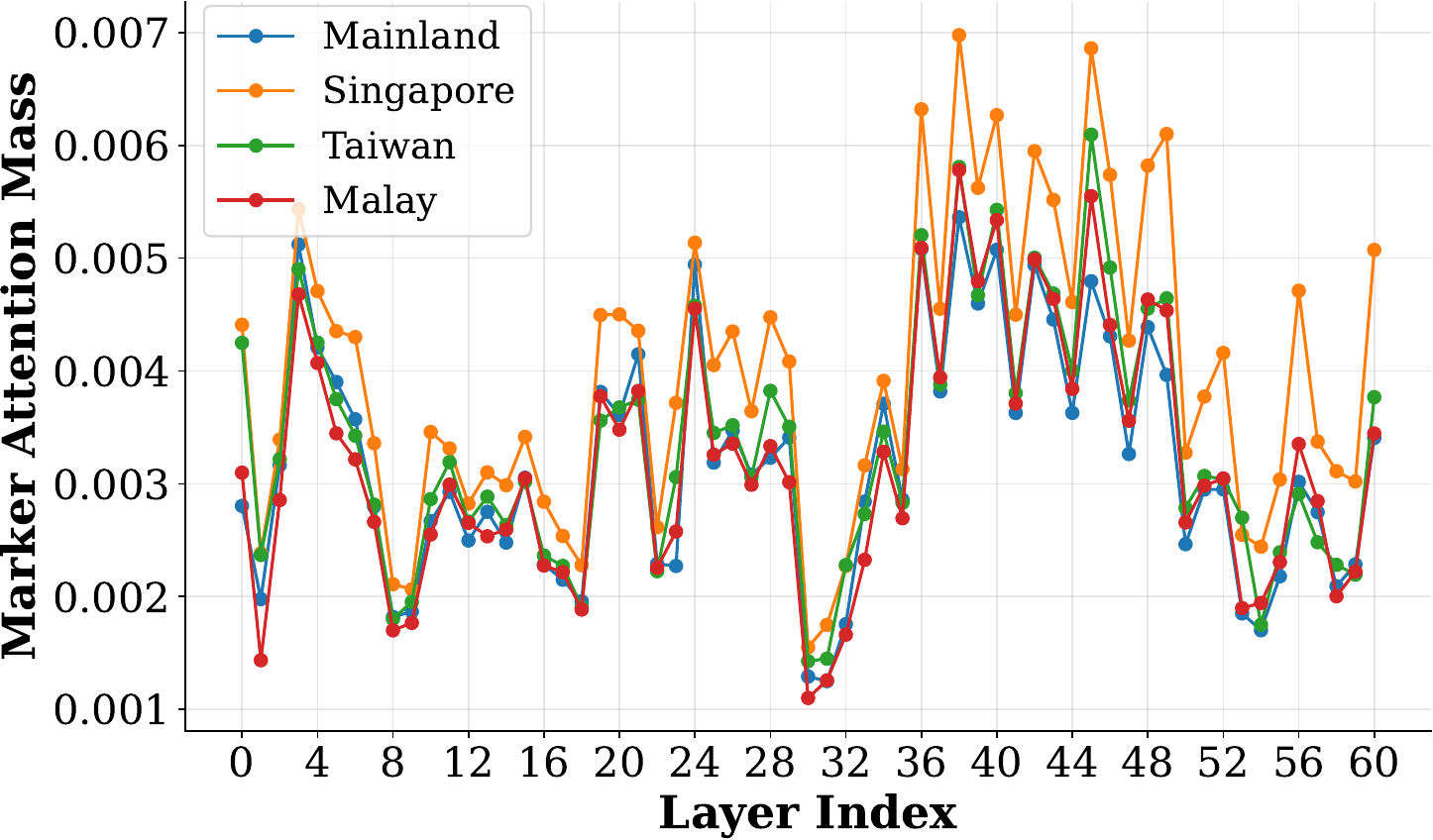}
  \caption{Layer-wise attention mass assigned to region-specific marker tokens across regional Chinese varieties.}
  \label{fig:Model attention by layers}
\end{figure}

Determining the loci of regional-understanding failures requires distinguishing an initial attention blind spot from a later semantic-integration difficulty. The layer-wise attention curves in Fig.~\ref{fig:Model attention by layers} provide stronger support for the latter, although attention alone does not constitute direct evidence of successful grounding. Because the Hong Kong items are largely written in Cantonese, including them in this diagnostic would conflate variety-level divergence with regional lexical variation. We therefore restrict the analysis to the other four regions. This exclusion is analysis-specific rather than benchmark-level: Hong Kong remains part of the main evaluation and is examined separately in~\ref{sec:Scope of Hong Kong inclusion across analyses}.

Table~3 summarizes four complementary diagnostics used to distinguish attention, answer confidence, terminology understanding, and grounding failure. Gold-answer probability (GoldP) is the probability assigned to the exact gold-answer combination under the forced-choice candidate-scoring procedure. We report terminology-explanation accuracy separately for items with correct final answers ($E_{\checkmark}$) and those with incorrect final answers ($E_{\times}$), with $\Delta = E_{\checkmark} - E_{\times}$ denoting the difference between the two in percentage points. Grounding-failure share ($G$) is the proportion of incorrect-answer cases that satisfy our operational criterion for grounding failure. Together, these measures allow us to assess whether attention to regional markers is accompanied by correct lexical interpretation and successful integration into the final prediction.

In the deeper layers, particularly layers 36--48, marker tokens from all four analyzed regions attract substantial attention, suggesting that the model does not simply overlook region-specific cues. However, this attention does not reliably translate into correct predictions. Singapore provides the clearest example: despite having the highest final-layer marker attention (0.0051), it exhibits the lowest GoldP (0.0747), a low $E_{\times}$ (40.0\%), and a high grounding-failure share ($G=0.60$). This combination indicates that many errors involve failures to interpret regional terminology correctly. The model may therefore register regionally marked tokens without reliably mapping them to locally licensed meanings. By contrast, Taiwan achieves the highest GoldP and an $E_{\checkmark}$ of 84.4\% despite receiving lower marker attention. A similar pattern appears for Malaysia, suggesting more efficient grounding of familiar or better-resourced regional varieties.

The Mainland data point to an additional compositional-integration bottleneck. Although $E_{\checkmark}$ is high at 81.9\%, Mainland items contain the largest average number of marker terms per instance (2.9231), while GoldP remains limited at 0.1625. Moreover, the relatively high $E_{\times}$ of 50.6\% shows that the model often explains individual terms correctly even when its final answer is wrong. The resulting gap between terminology-level understanding and answer-level performance suggests that the model may recognize isolated markers but fail to combine multiple localized cues into the regionally licensed interpretation. These results point to lexical disambiguation and compositional-propagation errors. Further details of the terminology-explanation experiment are provided in~\ref{sec:model terminology explanation experiment}.

\begin{table}[t]
\centering
\small
\setlength{\tabcolsep}{3.5pt}
\renewcommand{\arraystretch}{0.9} 
\begin{tabular}{lrrrrrrr}
\toprule
Region & Attn. & GoldP & T/S
  & $E_{\checkmark}$ & $E_{\times}$ & $\Delta$ & $G$ \\
\midrule
Mainland  & 0.0034 & 0.163 & 2.92 & 81.9 & 50.6 & 31.4 & 0.49 \\
Singapore & 0.0051 & 0.075 & 1.62 & 69.3 & 40.0 & 29.3 & 0.60 \\
Taiwan    & 0.0038 & 0.341 & 1.64 & 84.4 & 59.3 & 25.1 & 0.41 \\
Malaysia  & 0.0034 & 0.272 & 1.51 & 76.1 & 38.6 & 37.4 & 0.61 \\
\bottomrule
\end{tabular}
\caption{Marker attention, gold-answer probability, density, and
term-explanation accuracy conditioned on answer correctness. $E_{\checkmark}$/$E_{\times}$: explanation accuracy
on correct/incorrect items; $\Delta = E_{\checkmark} - E_{\times}$
(pp); $G$: grounding-failure share among wrong-answer items,
$d/(c+d)$.}
\label{tab:attn-grounding}
\end{table}

\subsection{Forced-Choice Calibration across Regions}
\label{sec:Confidence without Understanding}

A forced-choice calibration analysis using GLM examines whether regional comprehension failures coincide with miscalibrated confidence. For each multiple-choice item, we score all valid answer combinations, including single options, option pairs, and all-choice candidates, and convert conditional likelihoods into predictive probabilities using temperature-scaled softmax ($T \in [0.5, 2.0]$; full setup in Section~\ref{sec: Results of Qwen and SeaLLMs}). Calibration is assessed with Expected Calibration Error (ECE)~\citep{guo2017calibration}, multiclass Brier score~\citep{glenn1950verification}, and Negative Log-Likelihood (NLL).

Increasing $T$ consistently reduces ECE, Brier score, and NLL across regions, while accuracy remains unchanged because temperature scaling preserves candidate rankings. The improvement therefore reflects reduced overconfidence, not better regional understanding. Diagnostically, a calibrated model should express uncertainty when localized lexical cues are unresolved; GLM instead often assigns high confidence to incorrect candidates, treating uncertain regional interpretations as settled choices.

Calibration difficulty is strongly region dependent. Singapore shows the worst miscalibration, with ECE $0.66$, Brier score $1.35$, and NLL $4.12$ at $T=1.0$; even at $T=2.0$, its ECE remains high at $0.42$. Mainland and Malaysia are intermediate, Taiwan is better calibrated, and Hong Kong improves most, with ECE dropping from $0.21$ to $0.07$. These results show that regional failures often appear as high confidence without understanding: temperature scaling reduces overconfidence but does not resolve errors in regional identification, lexical interpretation, or compositional reasoning. This complements the attention analysis, showing that models may attend to regional markers yet still make confident forced-choice errors when local meanings are not integrated into the final decision.

\begin{figure}[t]
  \centering
  \captionsetup{skip=0pt}
  \includegraphics[width=0.43\textwidth]{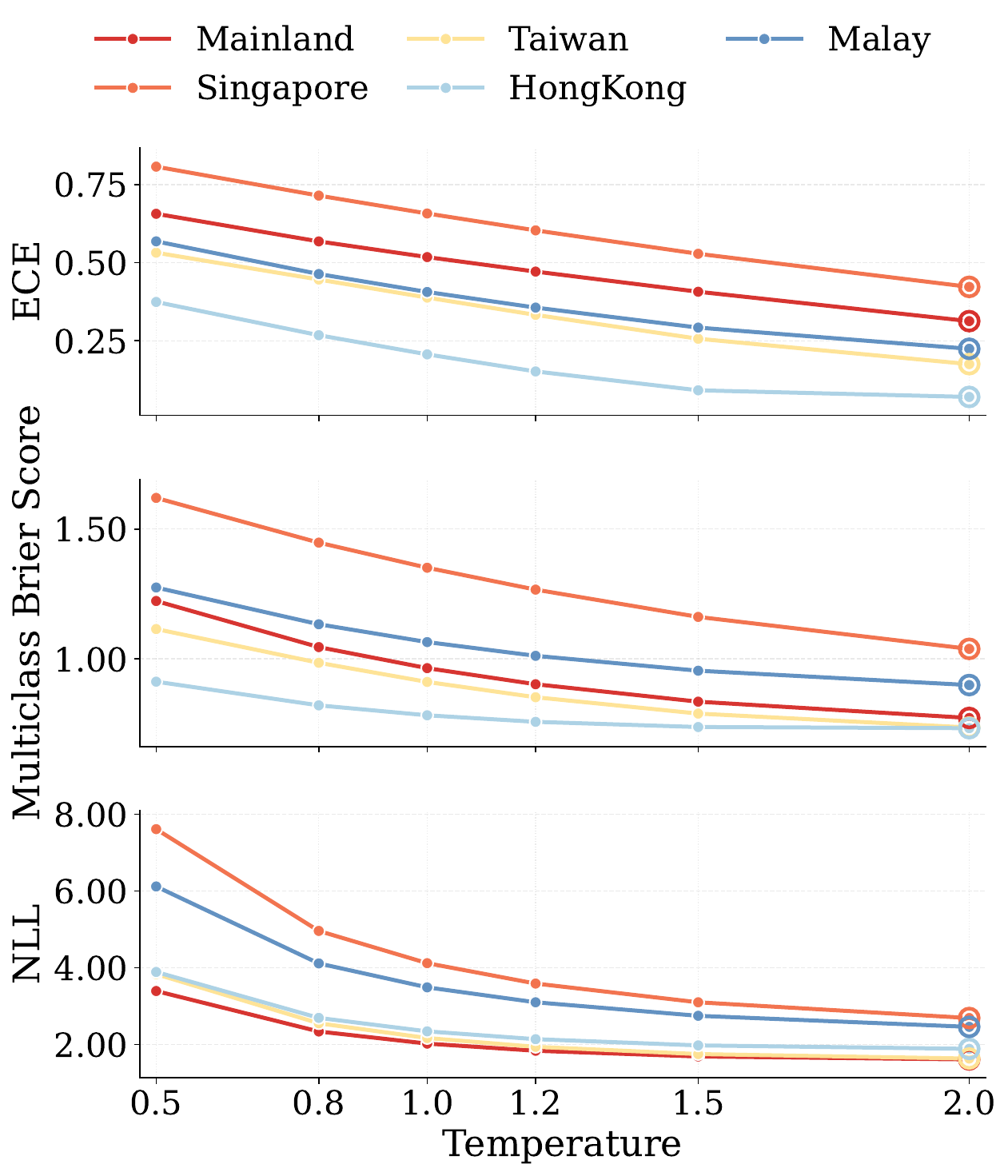}
  \caption{Temperature-scaled decoding calibration using forced-choice candidate likelihoods. Exact values for all data points are provided in~\ref{sec: Results of Qwen and SeaLLMs}.}
  \label{fig:all}
\end{figure}

\subsection{Surface Feature Interference across Regions and Domains}
\label{sec: Surface Feature Interference}

\begin{figure}[t]
  \centering
  \captionsetup{skip=0pt}
  \includegraphics[ width=0.49\textwidth]{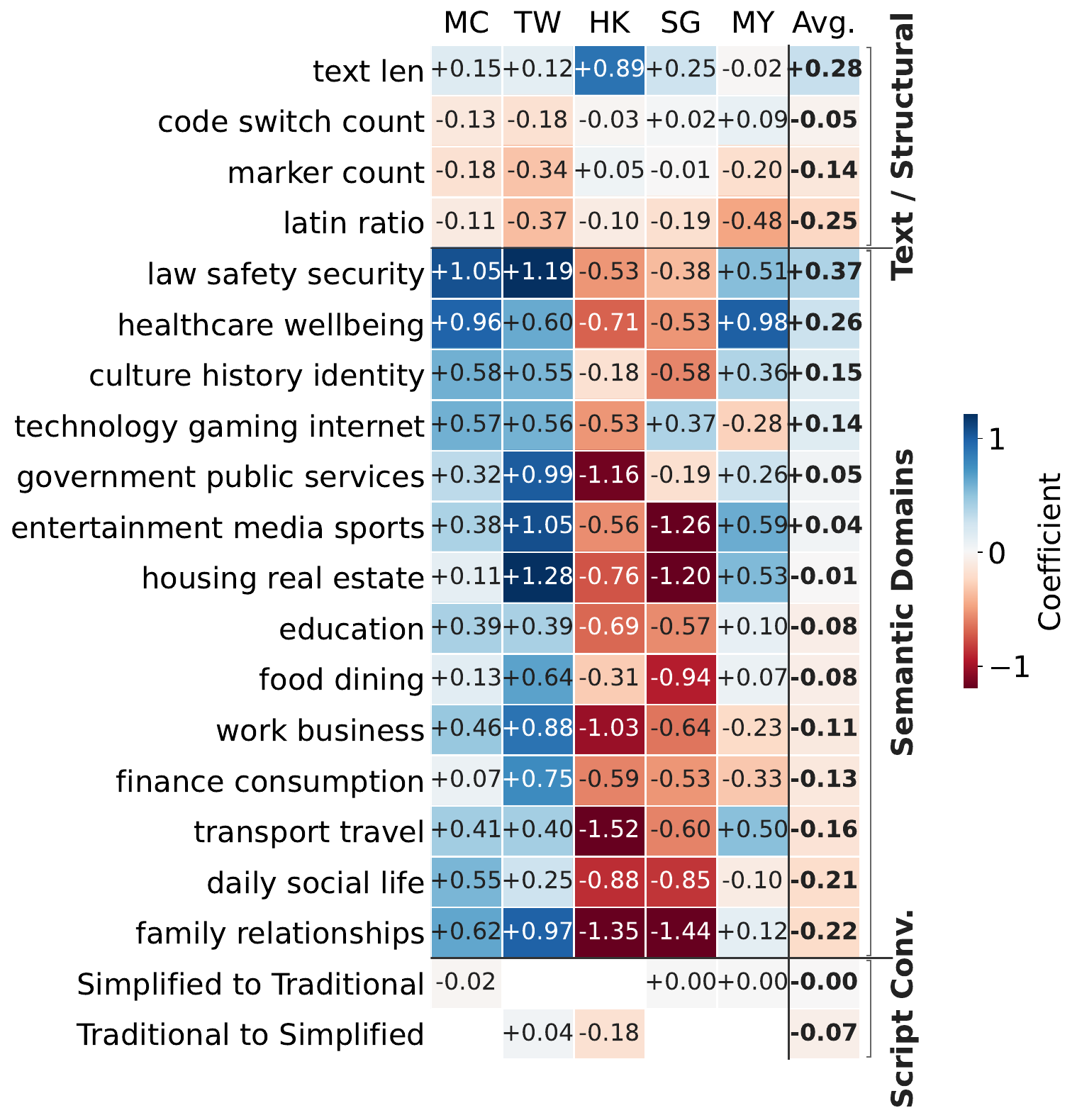}
  \caption{Regression coefficients of text features, semantic domains, and script conversion effects.}
  \label{fig:Grouped Coefficient Plot}
\end{figure}

We conduct a pooled regression across three representative models to identify surface features associated with regional Chinese understanding. As shown in Fig.~\ref{fig:Grouped Coefficient Plot}, marker count and Latin character ratio correlate most strongly with lower performance, with coefficients of $-0.14$ and $-0.25$. This suggests that errors are linked to dense localized cues and surface forms requiring regional grounding. Marker count is more than a length effect, as each localized marker adds a regional constraint to be jointly interpreted; by contrast, longer text correlates positively with performance, suggesting that richer context supports reasoning.

The effect of code-switching is region dependent. In Singapore and Malaysia, it correlates positively with performance, likely helping regional identification. In Mainland China and Taiwan, where such mixing is less typical, it correlates negatively ($-0.13$ and $-0.18$), showing that the same surface feature can aid or hinder interpretation depending on local usage patterns.

Script conversion is also region sensitive. Although aggregate effects are small, converting Hong Kong inputs from Traditional to Simplified Chinese substantially reduces performance, while Taiwan shows a slightly positive effect. This suggests that Hong Kong Traditional Chinese carries intertwined script, lexical, contact-induced, and syntactic cues, whereas Taiwan's relevant lexical meanings are more robust across scripts.

Semantic domain further modulates difficulty. Models are more robust in standardized domains such as law, safety, security, healthcare, government services, and cultural history, but weaker in everyday domains such as food, social life, education, business, transport, and finance. Overall, RegioZH failures reflect localized lexical burden, region-dependent script signals, and domain-specific language, reinforcing its role as a benchmark for region-licensed interpretation rather than general Chinese fluency or simple dialect robustness.

\subsection{Synthesizing the Diagnostics: A Regional Reasoning Failure Pathway} 
\label{sec:regional_failure_pathway} 

The four diagnostics, though correlational, point toward a shared failure pathway rather than four unrelated phenomena. Regional surface properties are associated with interpretive difficulty. Dense localized markers, region-dependent code-switching, script conventions, and everyday domains predict lower accuracy, plausibly by increasing the number or ambiguity of region-specific constraints to be resolved. Some of these forms are also encoded apart from standardized Chinese, particularly when lexical, syntactic, and script variation co-occur. Representational separation alone does not track failure, however: most regional inputs converge after formalization and script unification, yet errors persist. The most consistent association arises at the mapping from detected cues to region-licensed meanings. Substantial marker attention does not reliably co-occur with correct interpretation; Singapore combines the highest marker attention with the lowest gold-answer probability and the largest grounding-failure share. Composition of multiple localized constraints shows a second pattern: on Mainland items, individual terms are often explained correctly even when the final answer is wrong, and marker density predicts lower accuracy. The errors are thus compatible with both unstable local grounding and incomplete propagation of recognized meanings into the final interpretation; our diagnostics do not separate their relative contributions. Forced-choice decoding co-occurs with, and may amplify, these unresolved failures. Models assign high probability to incorrect candidates when regional meanings remain ambiguous, and temperature scaling reduces overconfidence without changing accuracy, suggesting miscalibration follows the interpretation failure instead of driving it. This synthesis also situates the prompting results in Sec.~5.2: prompt changes may perturb regional salience, answer priors, parsing behavior, or knowledge access, but correlational gains do not identify which. We present this layered account as a working hypothesis that organizes the observed associations and motivates targeted interventions.

\section{Error Attribution and Qualitative Analysis}

\begin{table}[t]
\small
\centering
\scriptsize
\setlength{\tabcolsep}{4pt}
\renewcommand{\arraystretch}{0.9}
\resizebox{\columnwidth}{!}{
\begin{tabular}{lcccc}
\toprule
\rowcolor{gray!12}
\textbf{Model} 
& \textbf{Error Rate} 
& \multicolumn{3}{c}{\textbf{Error Type Distribution}} \\
\rowcolor{gray!12}
& \textbf{(\%)} 
& \textbf{Term.} 
& \textbf{Socio.} 
& \textbf{Other} \\
\midrule
SeaLLMs-v3-7B-Chat & 58.48 & 67.46 & 17.74 & 14.80 \\
GLM-4-32B-0414     & 40.73 & 52.30 & 26.25 & 21.45 \\
\textbf{Qwen3-30B-A3B} 
& \textbf{38.58} & 59.05 & 30.71 & 10.24 \\
\bottomrule
\end{tabular}
}
\caption{Error attribution across models. Errors are categorized into regional terminology understanding errors, sociocultural inference errors, and other types.}
\label{tab:error_analysis}
\end{table}

\begin{figure*}[t!]
  \centering
  \captionsetup{skip=0pt}
  \includegraphics[width=\textwidth]{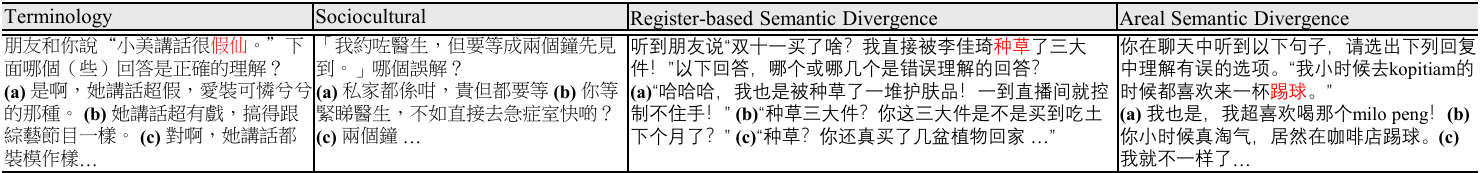}
  \caption{Examples of understanding errors across different semantic categories (English counterparts  are in Fig.~\ref{fig:regional-error-examples-english}).}
  \label{fig:regional-error-examples}
\end{figure*}

\begin{CJK}{UTF8}{gbsn}
Building on the preceding diagnostics, we conduct targeted attribution and case studies of errors on the RegioZH task: failures to select the region-licensed interpretation of localized Chinese expressions. We classify each incorrect prediction into three categories: regional terminology understanding, sociocultural inference, and other errors, using an LLM-as-judge protocol. Human--model agreement is reported in~\ref{sec:H-M agreement}. These labels are intended to localize the apparent source of a region-licensed interpretation failure, rather than to be read as direct evidence about a model's internal mechanism.


\subsection{Terminology and Sociocultural Errors across Model Capability Tiers} 

Tab.~\ref{tab:error_analysis} shows that RegioZH errors concentrate in two region-sensitive categories: terminology understanding and sociocultural inference. Terminology errors are the largest share across models, reaching 67.46\% for SeaLLMs, indicating that region-specific lexical items are a major failure source. For example, the Taiwanese expression ``假仙'' in Fig.~\ref{fig:regional-error-examples} requires a local interpretation; otherwise, models may choose an option based on a literal or nonlocal reading.

Error composition also differs across models. Although Qwen and GLM make fewer total errors than SeaLLMs, sociocultural errors form larger shares of their remaining mistakes: 30.71\% and 26.25\%, compared to 17.74\% for SeaLLMs. Thus, lower overall error rates do not eliminate region-sensitive failures; remaining errors often involve local institutions, social practices, and shared assumptions. The Hong Kong medical appointment case in Fig.~\ref{fig:regional-error-examples}, for instance, requires implicit knowledge of public/private healthcare, emergency wait times, and visitation expectations rather than special terminology. Reliable inference therefore requires situating utterances in social and institutional context, not merely recognizing localized words.



\subsection{Semantic Divergence across Registers and Regions}

Terminology errors reflect semantic divergence, not just missing vocabulary. We distinguish register-based divergence, where a lexical item gains an informal sense in a specific communicative setting, from areal divergence, where meanings shift across regions. Fig.~\ref{fig:regional-error-examples} illustrates both: in Mainland Chinese, ``种草'' literally means planting grass but informally means developing interest through recommendation; in Malaysian and Singaporean contexts, ``踢球'' can refer not to playing football but to Milo, a chocolate malt drink. Relying only on the literal meaning or on one regional sense can therefore lead to mismatched interpretations. Additional cases are in~\ref{sec:error case study}.

These findings show that RegioZH targets a weakness often hidden by broad Chinese-language evaluations: models may produce fluent Chinese yet select meanings incompatible with the item's regional frame. RegioZH should therefore be interpreted as a diagnostic of region-licensed lexical reasoning, not general Chinese proficiency or dialect robustness. The attribution results also leave open whether errors stem mainly from uneven pretraining coverage, post-training bias toward majority-region senses, weak use of regional cues, or forced-choice overconfidence. Future work can test these alternatives through controlled data interventions, cue ablations, retrieval or grounding methods, and region-specific calibration analyses.

Overall, regional Chinese failures are not isolated vocabulary gaps; they involve local institutional knowledge, shared social assumptions, and semantic shifts across registers and regions. RegioZH thus provides a diagnostic resource for studying how LLMs map surface expressions to sociolinguistic, cultural, and institutional contexts across Chinese-speaking regions.


\end{CJK}


\section{Conclusion}

This work challenges the view of Chinese proficiency in LLMs as monolithic by introducing RegioZH, a 2,607-item benchmark for regional lexical reasoning across Mainland China, Taiwan, Hong Kong, Singapore, and Malaysia. RegioZH targets region-licensed interpretation of localized Chinese expressions, evaluating whether models select meanings grounded in local lexical, sociocultural, and institutional usage. Our evaluation reveals substantial cross-regional disparities. Diagnostic analyses show that models may notice localized markers yet still fail to select region-appropriate meanings, often with overconfident forced-choice errors. These findings highlight intra-language regional interpretation as a key LLM evaluation dimension. Future work can broaden RegioZH across communities, domains, and evolving expressions, while extending evaluation beyond MCQs and developing methods that improve regional interpretation without homogenizing local variation.

\section*{Limitations}

This work primarily uses Chinese as a testbed to examine disparities in models’ semantic reasoning abilities across regional variants of the same language. It does not yet cover regional variants in other languages. We plan to extend this investigation to a broader range of languages in future work.  Additionally, \textsc{RegioZH} is currently constructed based on a multiple-choice format. While this design choice facilitates standardized automated evaluation and systematic decoding calibration analysis, it primarily assesses the models' discriminative capabilities. Model performance and regional comprehension barriers might exhibit different dynamics in free-form generative tasks, such as multi-turn dialogues or open-ended question answering. We plan to develop generative evaluation protocols in future extensions to provide a more comprehensive assessment. Current improvements in reasoning over regional Chinese variants rely primarily on prompting, with limited gains.Future work will explore reinforcement learning~\citep{li2026draftrl,li2026boostapr,tan2026evocurl}, retrieval-augmented methods~\citep{li2026retrieval, yu2026detectingresolvingmonitoringcontrol,yu2026fidesfaithfulinferencedeep}, or knowledge distillation~\citep{zheng2026trajectory,zheng2026pilot} to further enhance model performance.

\section*{Acknowledgments}

This research project is supported by the National Research Foundation, Singapore, under its National Large Language Models Funding Initiative (AISG Award No: AISG-NMLP-2024-003, AISG Award No: AISG-NMLP-2024-004). Any opinions, findings and conclusions or recommendations expressed in this material are those of the author(s) and do not reflect the views of the National Research Foundation. We would also like to express sincere gratitude to Microsoft Research's Agentic AI Research and Innovation (AARI) programme for their financial and computing support.

This research is supported by the National Research Foundation, Singapore under its National Large Language Models Funding Initiative. Any opinions, findings, conclusions, or recommendations expressed in this material are those of the author(s) and do not reflect the views of the National Research Foundation, Singapore.

\section*{Ethics Statement}

\paragraph{Bias and Representation:} \textsf{RegioZH} is designed to reveal region-dependent performance gaps of LLMs across regional varieties of Chinese, including disparities shaped by uneven coverage of Mainland China, Taiwan, Hong Kong, Singapore, and Malaysia in pretraining corpora and by imbalanced exposure to localized idioms, slang, code-mixing, politeness conventions, and sociocultural expressions. At the same time, \textsf{RegioZH} may still reflect residual biases from the linguistic intuitions of individual annotators (e.g., over-representation of urban or generationally specific usage), from reference materials such as Wikipedia, Baidu Baike, and regional dictionaries, or from question framing; some region-grounded items could also inadvertently reinforce stereotypes if not carefully curated. To mitigate these risks, annotators were required to verify each region-specific expression against authoritative sources, cross-check ambiguous cases against multiple independent references, and balance semantic-domain coverage. Three Chinese-linguistics experts then independently audited every item to correct errors, prune low-quality samples, adjudicate borderline cases, and remove content that could be misleading or stereotyping, achieving substantial inter-reviewer agreement (Fleiss' $\kappa = 0.762$).

\paragraph{Data Provenance and Annotation:} \textsf{RegioZH} is collaboratively constructed by region-specific expert teams. Annotators are native speakers of the regional variety they cover (Mainland China, Hong Kong, Taiwan, Singapore, or Malaysia) and were recruited based on long-term local residence and demonstrated familiarity with locale-bound lexical, pragmatic, and sociocultural conventions. All annotators received a project-wide briefing, participated in focus-group discussions to define semantic domains and question types, and followed detailed annotation guidelines covering category balance, source verification, distractor design, and sociocultural appropriateness. For data sourcing, items are authored from scratch rather than scraped, and annotators cross-checked region-specific meanings, usage contexts, and sociocultural assumptions against multiple authoritative references (Wikipedia, Baidu Baike, regional dictionaries, news articles, and other locally grounded materials). Distractors are required to be semantically plausible and comparable to the correct answer in form and specificity, so that performance reflects genuine regional competence rather than surface-level pattern matching. All items are authored in the appropriate regional written form, with English glosses produced and manually verified for documentation purposes. Annotators were compensated at competitive local rates (see Table~\ref{tab:annotator_info}), and no personally identifiable information about annotators is disclosed. We additionally filter content that could constitute hate speech, defamation, or sensitive personal information, and avoid items that depend on private or identifying details of real individuals.

\paragraph{Intended Benefit:} We aim to provide a transparent benchmark and diagnostic protocol that helps surface intra-language disparities, fragile regional grounding, and confidence miscalibration in Chinese-capable LLMs, rather than merely ranking models on aggregate accuracy. By pairing multiple-choice items with region-grounded interpretations, requiring models to explain their reasoning over locale-bound lexical cues, and complementing accuracy with representation, attention, calibration, and surface-feature analyses, the benchmark encourages region-sensitive grounding and discourages shortcut solving based on standardized Mandarin priors. We will release \textsf{RegioZH} and supporting resources for research use to facilitate reproducibility, bias and fairness analysis, and future work that broadens high-quality coverage for under-served regional varieties of Chinese and other languages with comparable intra-language heterogeneity.



\bibliography{custom}

\appendix


\section{Extended Related Work}
\label{sec:related work}
\textbf{Cultural alignment and knowledge elicitation.} Benchmarks such as CDEval~\citep{wang2024cdeval}, MMA-ASIA~\citep{zheng2025mma}, and BLEnD-Vis~\citep{tan2026blend} evaluate cultural knowledge and geo-cultural dimensions at a broad level. Our work targets a more grounded capability: the interpretation of localized everyday expressions where failures arise from lexical semantic drift across regions or from missing socially embedded institutional knowledge. Our finding that persona prompting and structured chain-of-thought partially recover regional performance further suggests that many failures reflect suboptimal knowledge elicitation rather than pure knowledge absence. This result complements cultural alignment work on prompting-based intervention.

\section{Benchmark Construction Details}

\label{sec:dataset creation}
During dataset construction, annotators were instructed to verify the meanings and contextual conditions of region-specific expressions, colloquialisms, and slang through authoritative reference sources, including Wikipedia, Baidu Baike, dictionaries, news articles, and other locally grounded materials. This step was designed to minimize ambiguity in lexical interpretation and to ensure that each expression was used appropriately within its intended sociocultural and pragmatic context.

For distractor design, annotators were required to construct alternatives that were semantically plausible and comparable to the correct answer in form, specificity, and topical relevance. We explicitly discouraged distractors that could be ruled out through superficial cues, such as obvious semantic mismatch, unnatural phrasing, or contextual implausibility. This design principle ensures that model performance reflects genuine regional semantic competence rather than shallow pattern matching or process-of-elimination strategies. Full annotation guidelines are provided in Table~\ref{tab:annotation_guideline}.

Table~\ref{tab:semantic_domain_distribution} presents the distribution of semantic domains across regions, and Table~\ref{tab:annotator_info} summarizes annotator recruitment details. For each target region, we recruited three annotators familiar with the corresponding regional variety of Chinese. The subsequent expert validation stage was conducted by a separate panel of three Chinese-linguistics reviewers. To protect annotator privacy, no personally identifiable information is disclosed. We report only annotator regions and compensation.

To assess the reliability of the expert validation stage, we computed Fleiss' $\kappa$ over all 2,750 items submitted for quality control, evaluated by three Chinese-linguistics reviewers. Fleiss' $\kappa$ is preferred over Cohen's $\kappa$, which is defined only for two annotators. The observed agreement rate was $\bar{P} = 0.881$ and the expected chance agreement was $P_e = 0.500$, yielding:
\begin{equation}
    \kappa = \frac{\bar{P} - P_e}{1 - P_e} = 0.762
\end{equation}
Under the \citet{landis1977measurement} scale, this falls within the \textit{substantial agreement} range ($0.61$--$0.80$), approaching the \textit{almost perfect} threshold. These results indicate substantial agreement among reviewers on the pre-adjudication three-way validation label space, namely retain, revise, or remove. Items for which a majority of reviewers voted for rejection were discarded, while those requiring revision were returned to the respective regional annotators. This process yielded the final benchmark of 2,607 items.

\begin{table}[htpb]
\centering
\begin{tabular}{lll}
\toprule
Annotator & Region & Compensation \\
\midrule
A & Mainland China & 1200 SGD/Month \\
B & Mainland China & Free \\
C & Mainland China & Free \\
D & Hong Kong & 1 SGD/Sample \\
E & Hong Kong & 1200 SGD/Month \\
F & Hong Kong & Free \\
G & Taiwan & 1200 SGD/Month \\
H & Taiwan & 1200 SGD/Month \\
I & Taiwan & 1200 SGD/Month \\
J & Singapore & 1200 SGD/Month \\
K & Singapore & 1200 SGD/Month \\
L & Singapore & Free \\
M & Malaysia & 1200 SGD/Month \\
N & Malaysia & Free \\
O & Malaysia & Free \\
\bottomrule
\end{tabular}
\caption{Annotator information.}
\label{tab:annotator_info}
\end{table}

\begin{table*}[t]
\centering
\small

\begin{tcolorbox}[colback=blue!3, colframe=black!60,
  title=\textbf{Stage 1: Category Distribution}]
Review the semantic category distribution agreed upon during group discussion before you begin.
Maintain the target proportions throughout annotation and adjust your item selection periodically.
Do not over-annotate in any single category, as balanced coverage across categories is essential
for dataset diversity.
\end{tcolorbox}

\begin{tcolorbox}[colback=gray!5, colframe=black!60,
  title=\textbf{Stage 2: Expression Verification}]
Verify the meaning and usage conditions of every region-specific expression, colloquialism, or
slang term through at least one authoritative reference source, such as Wikipedia, Baidu Baike,
regional dictionaries, or news articles. 
\end{tcolorbox}

\begin{tcolorbox}[colback=gray!5, colframe=black!60,
  title=\textbf{Stage 3: Cross-Source Verification}]
If a term's meaning or usage remains unclear after consulting one source, verify it against at
least two independent sources of different types (e.g., a dictionary and a news article). If
sources conflict or sufficient evidence cannot be found, \textbf{do not use the expression}.
Do not rely on personal intuition. \textit{When in doubt, leave it out.}
\end{tcolorbox}

\begin{tcolorbox}[colback=yellow!5, colframe=black!60,
  title=\textbf{Stage 4: Item Construction}]
Each item must include: (1) a context sentence in which the target expression appears naturally,
(2) a question probing its meaning or pragmatic implication, (3) at least one correct answer grounded in
verified sources, and (4) a reasonable number of distractors. Avoid contexts that make the correct answer
identifiable without genuine regional knowledge.
\end{tcolorbox}

\begin{tcolorbox}[colback=yellow!5, colframe=black!60,
  title=\textbf{Stage 5: Distractor Design}]
Distractors must be semantically plausible and comparable to the correct answer in form,
specificity, and topical relevance. Avoid distractors eliminable through obvious semantic
mismatch, unnatural phrasing, contextual implausibility, or trivial surface-form differences
from the correct answer.
\end{tcolorbox}

\begin{tcolorbox}[colback=orange!5, colframe=black!60,
  title=\textbf{Stage 6: Sociocultural and Pragmatic Appropriateness}]
Place each expression within its appropriate sociocultural and pragmatic context, attending to
register, formality, and the regional community of use. If an expression is restricted to a
particular group or setting, provide sufficient contextual framing. Revise or discard items
that require knowledge beyond what a reasonably informed regional speaker would possess.
\end{tcolorbox}

\begin{tcolorbox}[colback=red!4, colframe=black!60,
  title=\textbf{Stage 7: Flagging, Review, and Revision}]
Flag any item with residual uncertainty regarding naturalness, regional specificity, or
distractor quality before submission. If an item is returned after expert review, revise it
according to reviewer comments or withdraw it if revision is not feasible. Consult your team
lead for recurring ambiguities not resolved by these guidelines.
\end{tcolorbox}
\caption{Annotation guidelines for dataset construction. Annotators are expected to follow
all stages described below in order, from pre-annotation planning through submission and revision.}
\label{tab:annotation_guideline}
\end{table*}

\begin{table*}[t!]
\centering
\setlength{\tabcolsep}{5pt}
\renewcommand{\arraystretch}{1.08}
\begin{tabular}{lrrrrr}
\toprule
\textbf{Semantic Domain} 
& \textbf{Hong Kong} 
& \textbf{Mainland China} 
& \textbf{Malaysia} 
& \textbf{Singapore} 
& \textbf{Taiwan} \\
\midrule
Culture, History, and Identity     & 20  & 16 & 58 & 37  & 24 \\
Daily Social Life                  & 79  & 83 & 30 & 80  & 78 \\
Education                          & 46  & 35 & 40 & 30  & 54 \\
Entertainment, Media, and Sports   & 54  & 63 & 31 & 22  & 41 \\
Family and Relationships           & 22  & 52 & 31 & 15  & 20 \\
Finance and Consumption            & 50  & 31 & 31 & 30  & 26 \\
Food and Dining                    & 45  & 35 & 45 & 96  & 45 \\
Government and Public Services     & 18  & 16 & 57 & 35  & 39 \\
Healthcare and Wellbeing           & 44  & 16 & 31 & 18  & 22 \\
Housing and Real Estate            & 19  & 20 & 31 & 25  & 17 \\
Law, Safety, and Security          & 15  & 20 & 36 & 18  & 25 \\
Technology, Gaming, and Internet   & 25  & 74 & 31 & 15  & 20 \\
Transport and Travel               & 19  & 16 & 31 & 45  & 43 \\
Work and Business                  & 64  & 43 & 31 & 49  & 81 \\
Other                              & 0   & 0  & 2  & 0   & 1  \\
\bottomrule
\end{tabular}
\caption{Distribution of semantic domains across regions.}
\label{tab:semantic_domain_distribution}
\end{table*}
\subsection{Dataset Examples}

\begin{figure*}[t!]
  \centering
  \captionsetup{skip=0pt}
  \includegraphics[width=\textwidth]{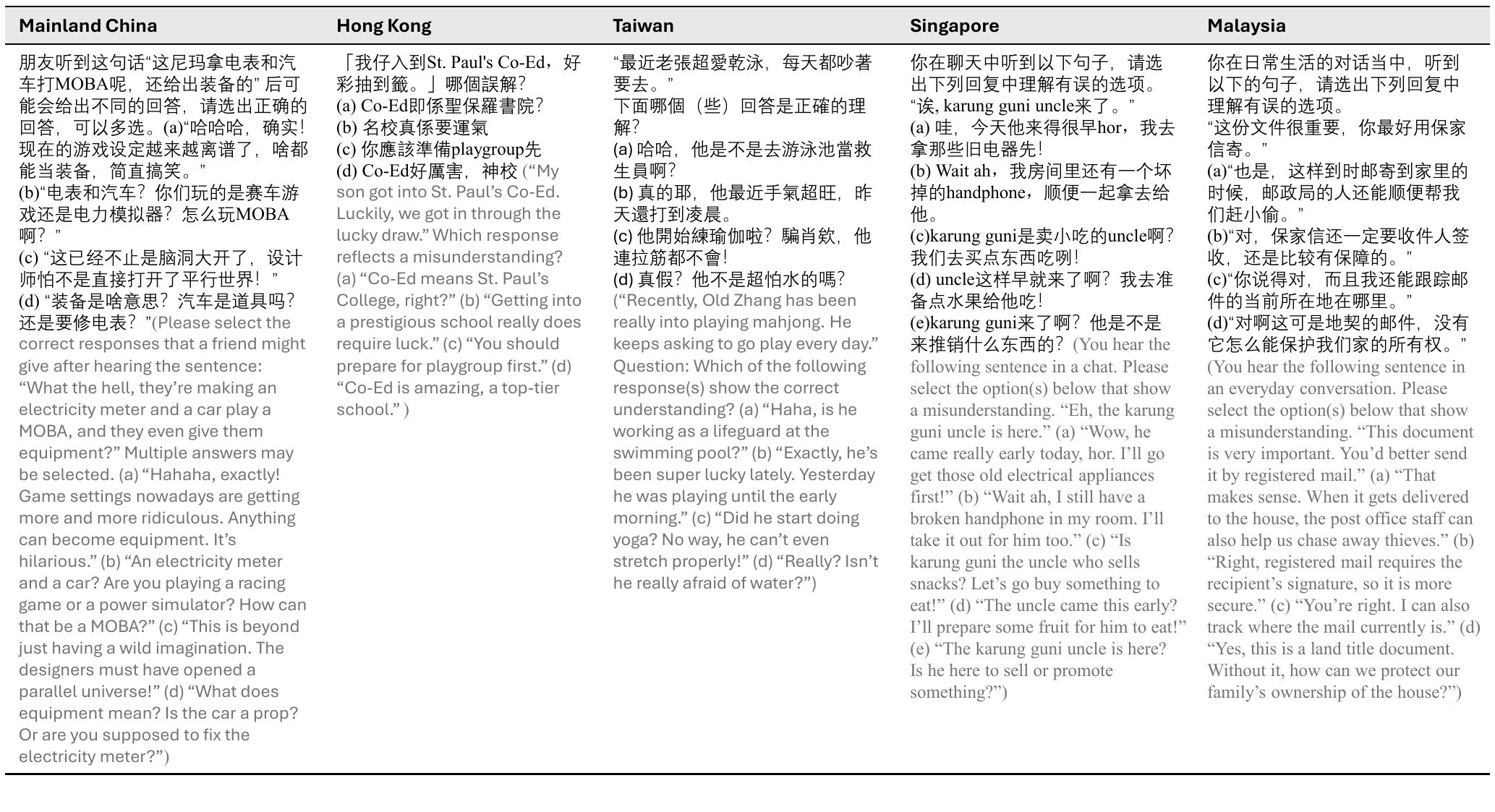}
  \caption{Dataset Examples}
  \label{fig:dataset examples}
\end{figure*}

We provide data samples from the five regions in the \textit{RegioZH} dataset in Fig.~\ref{fig:dataset examples}.

\section{Experimental Setup}

\subsection{Model Summary}
\label{sec:model_sum}

Table~\ref{tab:model_summary} provides a comprehensive summary of the models utilized in this study, including their respective categorizations and the specific experimental tasks to which each was applied.
\begin{table*}[t]
\centering
\footnotesize
\renewcommand{\arraystretch}{1.15}
\begin{tabularx}{\textwidth}{
>{\raggedright\arraybackslash}p{4cm}
>{\raggedright\arraybackslash}p{3cm}
>{\raggedright\arraybackslash}X
}
\toprule
\textbf{Model} & \textbf{Model Category} & \textbf{Experiments Used} \\
\midrule
DeepSeek-V3.1~\citep{liu2024deepseek} 
& Chinese-centric model 
& Zero-shot; Persona prompting; Structured CoT prompting \\

GPT-5.1 
& Global general-purpose model 
& Zero-shot; Persona prompting; Structured CoT prompting \\

Qwen3-30B~\citep{yang2025qwen3} 
& Chinese-centric model 
& Zero-shot; Persona prompting; Structured CoT prompting; Mechanistic analysis \\

ERNIE-4.5-21B~\citep{ernie2025technicalreport} 
& Chinese-centric model 
& Zero-shot \\

GLM-4-32B~\citep{glm2024chatglm} 
& Chinese-centric model 
& Zero-shot; Mechanistic analysis \\

Baichuan-M2-32B~\citep{dou2025baichuan} 
& Chinese-centric model 
& Zero-shot \\

SEA-LION-v4-27B~\citep{ng2025sea} 
& Southeast Asia-oriented model 
& Zero-shot \\

SeaLLMs-v3-7B~\citep{zhang2025seallms} 
& Southeast Asia-oriented model 
& Zero-shot; Mechanistic analysis \\

Llama-3.1-8B~\citep{grattafiori2024llama} 
& Global general-purpose model
& Zero-shot; Persona prompting; Structured CoT prompting  \\

Gemma-7B~\citep{team2024gemma} 
& Global general-purpose model 
& Zero-shot \\

Ministral-3-14B~\citep{liu2026ministral} 
& Global general-purpose model
& Zero-shot; Persona prompting; Structured CoT prompting  \\
\bottomrule
\end{tabularx}
\caption{Summary of models used in the experiments.}
\label{tab:model_summary}
\end{table*}




\begin{figure*}[t!]
  \centering
  \captionsetup{skip=0pt}
  \includegraphics[width=\textwidth]{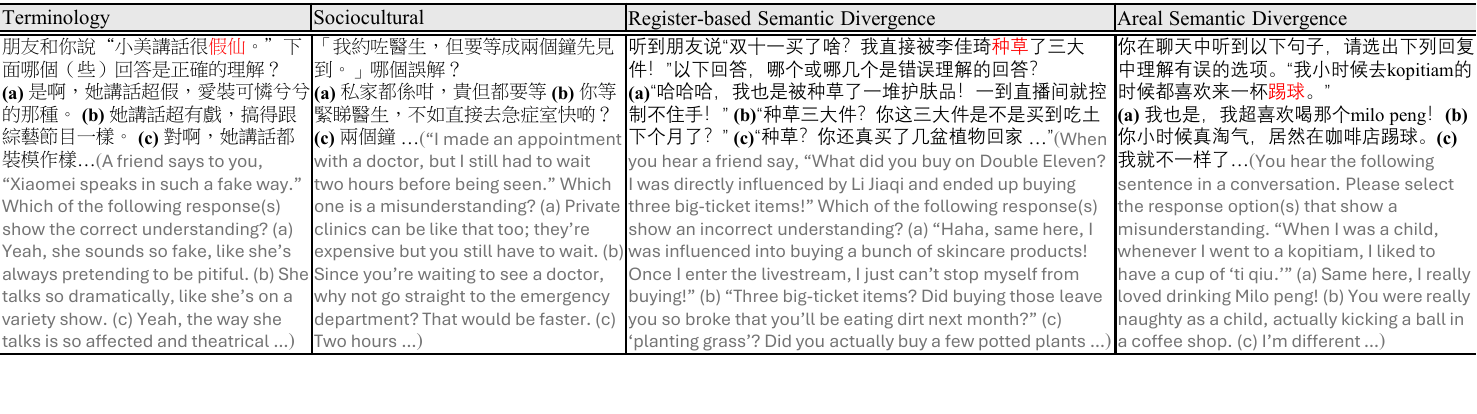}
  \caption{Examples of regional Chinese understanding errors across different semantic categories with English translations.}
  \label{fig:regional-error-examples-english}
\end{figure*}

\subsection{Prompting Strategy Evaluation}
\label{sec:Experiment setting and Prompt}

We conducted experiments under three prompting settings: zero shot prompting, persona based prompting, and structured chain of thought prompting, with Temperature = 1 and top\_p = 1. The specific prompt used for each strategy is provided in Table~\ref{tab:prompt-settings}.

\begin{table*}[t]
\centering
\small

\begin{tcolorbox}[colback=blue!3, colframe=black!60,
  title=\textbf{Zero-shot Prompt}]
Please answer the following question. Select the correct option(s). Note: The question
may be multiple-choice, i.e., more than one option can be correct.\\[4pt]

\texttt{\{question\_and\_answer\}}\\[4pt]

Please provide your response in the following format:\\[2pt]

\texttt{ANSWER: [letter(s) of correct option(s), e.g., ``a'' or ``a,c'']}
\end{tcolorbox}

\vspace{4pt}

\begin{tcolorbox}[colback=gray!5, colframe=black!60,
  title=\textbf{Persona Prompting}]
You are a Chinese linguistic and cultural expert specializing in the
\texttt{\{region\}} region. You have deep knowledge of regional Chinese language usage,
colloquialisms, cultural context, and pragmatic meaning in everyday conversations.
Please answer the following question. Select the correct option(s) and explain the
reason for your choice. Note: The question may be multiple-choice, i.e., more than
one option can be correct.\\[4pt]

\texttt{\{question\_and\_answer\}}\\[4pt]

Please provide your response in the following format:\\[2pt]

\texttt{ANSWER: [letter(s) of correct option(s), e.g., ``a'' or ``a,c'']}
\end{tcolorbox}

\vspace{4pt}

\begin{tcolorbox}[colback=yellow!5, colframe=black!60,
  title=\textbf{Structured CoT Prompting}]
You are evaluating a Chinese multiple-choice question based on regional linguistics.\\[4pt]

Analyze the \texttt{<question>} using the following criteria:

\begin{enumerate}\itemsep2pt
  \item \textbf{Region:} Identify the most likely source: Mainland China, Hong Kong,
        Taiwan, Singapore, or Malaysia.
  \item \textbf{Localized Terms:} Extract region-specific expressions, including
        dialects, slang, and transliterations. Map each to standard Chinese or briefly
        explain it. If none exist, output \texttt{- None}.
  \item \textbf{Answer:} Select all correct options based purely on linguistic meaning,
        not stereotypes.
\end{enumerate}

\texttt{<question>}\\
\texttt{\{question\_and\_answer\}}\\
\texttt{</question>}\\[4pt]

Output strictly using the exact template below. Ensure the final Answer contains only
lowercase letters, comma-separated with no spaces if multiple options are correct,
e.g., \texttt{\{\{a,c\}\}}.\\[4pt]

\texttt{Region: <region>}\\
\texttt{Localized Terms:}\\
\texttt{- <term> -> <standard equivalent OR explanation>}\\
\texttt{Answer: \{\{<options>\}\}}
\end{tcolorbox}
\caption{Prompts used in three prompting settings. The same question field
\texttt{\{question\_and\_answer\}} is filled at inference time. The persona prompt
additionally uses the target region field \texttt{\{region\}}.}
\label{tab:prompt-settings}
\end{table*}

\subsection{Scope of Hong Kong inclusion across analyses.}
\label{sec:Scope of Hong Kong inclusion across analyses}
Our four diagnostic lenses operate at different linguistic granularities, 
and we adopt analysis-specific inclusion criteria for Hong Kong accordingly. 
Representation geometry (Sec.~\ref{sec:Representation Geometry}), forced-choice calibration (Sec.~\ref{sec:Confidence without Understanding}), and 
surface-feature regression (Sec.~\ref{sec: Surface Feature Interference}) operate at the sentence or option level 
and are robust to lexical and orthographic divergence; including Hong Kong 
in these analyses is essential, as our goal is precisely to characterize 
its outlying behavior. Attention attribution (Sec.~\ref{sec:Attention without Grounding}), in contrast, requires 
identifying region-specific \textit{marker tokens} against a Mandarin-based 
written norm. Because the Hong Kong subset is predominantly written in 
Cantonese, which diverges from Mandarin at the lexical, syntactic, and 
orthographic levels, the very notion of an isolated regional marker becomes 
ill-defined: virtually every token could be classified as region-specific. 
Including Hong Kong would conflate variety-level linguistic divergence with 
the lexical-marker-level signal under investigation. We therefore restrict 
the attention analysis to the four regions whose written form remains 
within the Mandarin orthographic norm.

\section{Evaluation Methodology and Judging Protocols}

\subsection{Terminology Explanation Experiment}
\label{sec:model terminology explanation experiment}
To examine whether model performance on regional Chinese semantic reasoning is linked to the accuracy with which models interpret region specific lexical items, we further require models to justify their predictions after producing the selected option(s). Specifically, the explanation must ground the decision to select or reject each option in the meanings of region specific terms appearing in both the question and the answer choices. Table~\ref{tab:answer_explanation_error_classification} presents the prompt used for this experiment. All decoding settings are kept consistent with the other experiments, with Temperature set to 1 and top\_p set to 1. 
We evaluate the generated explanations using an LLM-as-Judge protocol. The judge focuses specifically on whether the model correctly interprets the region specific lexical items involved in its reasoning. An instance is marked as correct only if the explanation provides accurate interpretations for all region specific terms appearing in the question and answer choices. If any such term is misinterpreted, omitted, or used to support an incorrect semantic distinction, the instance is marked as incorrect. The judging prompt is provided in Table~\ref{tab:answer_explanation_error_classification}. In Section~\ref{sec:H-M agreement}, we further report a human model agreement study, which shows that the adopted multi model voting protocol achieves high judgment reliability while mitigating biases introduced by any single judge model.

\begin{table}[t!]
\centering
\setlength{\tabcolsep}{4pt}
\renewcommand{\arraystretch}{1.05}
\begin{tabular}{lcc}
\toprule
\textbf{Model} & \textbf{EC} & \textbf{TE} \\
\midrule
Claude-haiku-4.5    & 76\% & 78\% \\
Gemini-3.1-flash   & 76\% & 73\% \\
GPT-5.4            & 83\% & 85\% \\
Multi-model voting & 84\% & 87\% \\
\bottomrule
\end{tabular}
\caption{Human--model agreement ratios. EC denotes Error Classification, and TE denotes Terminology Explanation.}
\label{tab:hm_agreement}
\end{table}

\tcbset{
  promptbox/.style={
    colframe=black!60,
    boxrule=0.5pt,
    arc=2pt,
    left=4pt,
    right=4pt,
    top=1pt,
    bottom=1pt,
    boxsep=2pt,
    toptitle=1pt,
    bottomtitle=1pt,
    before skip=1pt,
    after skip=1pt
  }
}
\begin{table*}[t]
\centering
\footnotesize

\begin{tcolorbox}[promptbox, colback=blue!3,
  title=\textbf{Answer with Explanation Prompt}]
Please answer the following question. Select the correct option(s) and explain the reason for your choice. Note: The question may be multiple-choice, i.e., more than one option can be correct.\\[0pt]
\texttt{\{question\_and\_answer\}}\\[0pt]
In your explanation, if there are any region-specific words, localized expressions, dialectal terms, or culturally specific references in the question or options, explain their meanings in context and use them to support your reasoning.\\[0pt]
Please provide your response in the following format:\\[0pt]
\texttt{ANSWER: [letter(s) of correct option(s), e.g., ``a'' or ``a,c'']}\\
\texttt{EXPLANATION: [your reasoning for why these options are correct or incorrect]}
\end{tcolorbox}

\vspace{2pt}

\begin{tcolorbox}[promptbox, colback=gray!5,
  title=\textbf{Lexical Interpretation Correctness Judging Prompt}]
\textbf{System Prompt:}\\[0pt]
You are a strict evaluator of lexical interpretation in Chinese multiple-choice question answering explanations.\\[0pt]
Your task is to judge whether the explanation correctly interprets the meanings of key words, regional expressions, slang, code-switched expressions, local institutional terms, or culturally localized phrases that appear in the question.\\[0pt]
\textbf{Judging rules:}
\begin{enumerate}[nosep,leftmargin=1.4em]
  \item Do not use or infer any hidden gold answer. Judge only from the visible question/options and the explanation text.
  \item Do not evaluate whether the model selected the correct option. The answer section has been removed intentionally.
  \item Return true only if the explanation's explicit lexical interpretations, glosses, paraphrases, or semantic claims about key expressions are correct.
  \item Return false if the explanation misinterprets any key term, regional expression, slang, code-switched expression, local institution, or culturally localized phrase.
  \item Return false if the explanation invents a meaning that is not supported by the question text or gives a misleading lexical gloss.
  \item If the explanation contains no meaningful lexical interpretation or term-level semantic analysis, return false.
  \item Minor omissions are acceptable if the lexical interpretations that are present are correct and sufficient for the relevant phrase being discussed.
\end{enumerate}
\textbf{Output strictly as JSON:}
{\ttfamily
\{
 ``Lexical\_Interpretation\_correctness'': true
\}
}or
{\ttfamily
\{
``Lexical\_Interpretation\_correctness'': false
\}
}\\[0pt]
\textbf{User Prompt:}\\[0pt]
Please judge whether the explanation correctly interprets the vocabulary and key expressions in the question.\\[0pt]
The model's answer section has been removed. Do not judge answer correctness.\\
Do not assume or use any gold answer. Do not infer correctness from a hidden answer key.\\
Return JSON only. Do not output explanations.\\[0pt]
\textbf{Question:}
\texttt{\{question\_text\}}\\[0pt]
\textbf{Model explanation with the ANSWER section removed:}
\texttt{\{explanation\_text\}}
\end{tcolorbox}

\vspace{2pt}

\begin{tcolorbox}[promptbox, colback=yellow!5,
  title=\textbf{Error Classification Prompt}]
You are an expert annotator for regional Chinese multiple-choice QA errors.\\[0pt]
\textbf{Task:} Given a question, the gold answer, the model's wrong parsed answer, and the model's chain-of-thought / reasoning output, classify the primary cause of the error into exactly one of three categories:
\begin{enumerate}[
  itemsep=2pt,
  parsep=0pt,
  topsep=0pt,
  partopsep=0pt,
  leftmargin=1.4em
]
  \item \texttt{\{TERMINOLOGY\}}\\[-1pt]
  Use this when the model's reasoning shows that it misunderstood a regional lexical item, colloquial expression, loanword, slang, literal meaning, option wording, or phrase-level semantics.\\[-1pt]
  This includes wrong glosses, wrong paraphrases, confusing one local term with another, or treating a region-specific expression as a different word.

  \item \texttt{\{SOCIOCULTURAL\}}\\[-1pt]
  Use this when the model appears to understand the words literally but lacks the necessary regional cultural, institutional, historical, social, policy, food-practice, local-custom, geography, or everyday background knowledge.\\[-1pt]
  This includes missing how a local object, institution, social practice, place, policy, role, or cultural convention works.

  \item \texttt{\{OTHERS\}}\\[-1pt]
  Use this only when the wrong answer is not primarily caused by regional terminology misunderstanding or missing sociocultural background.\\[-1pt]
  This includes arithmetic mistakes, careless reasoning, logical inconsistency, hallucinated facts unrelated to regional Chinese, option-selection mistakes, incomplete reasoning, generic world-knowledge errors, parsing/formatting artifacts, or cases where the COT does not provide enough evidence to support either of the two regional-understanding categories.
\end{enumerate}
\textbf{Decision rule:}
\begin{itemize}[
  itemsep=0pt,
  parsep=0pt,
  topsep=0pt,
  partopsep=0pt,
  leftmargin=1.4em
]
  \item If both are present, choose the single most direct cause of the wrong answer according to the model's own reasoning.
  \item Use \texttt{\{OTHERS\}} for generic parsing or formatting errors unless the COT clearly indicates terminology misunderstanding or sociocultural background as the root cause.
  \item Prefer \texttt{\{OTHERS\}} over forcing a regional-understanding label when the evidence is ambiguous or insufficient.
  \item The model's answer is known to be wrong; do not re-solve the whole task unless needed to identify the failure mode.
\end{itemize}
Output only a JSON object with exactly these keys:\\[2pt]
{\ttfamily
\{``error\_type'': ``\{TERMINOLOGY\}'' or ``\{SOCIOCULTURAL\}'' or ``\{OTHERS\}'',\\
\quad ``confidence'': a number from 0 to 1,\\
\quad ``reason'': ``one concise sentence explaining the classification'',\\
\quad ``evidence'': ``a short phrase from the COT or question supporting the classification''
\}
}
\end{tcolorbox}
\caption{Prompts used for answer explanation generation, lexical interpretation
correctness judging, and error classification. Fields enclosed in
\texttt{\{...\}} are filled at inference time.}
\label{tab:answer_explanation_error_classification}
\end{table*}

\subsection{Error Type Classification Protocol}
\label{sec:Terminology and Sociocultural Errors}
We provide the prompt used to classify terminology understanding errors and sociocultural inference errors in Table~\ref{tab:answer_explanation_error_classification}. All decoding settings are kept consistent with the other experiments, with Temperature set to 1 and top\_p set to 1.

\subsection{Human Model agreement analysis}
\label{sec:H-M agreement}

To assess the reliability of our LLM-as-Judge framework, we randomly sampled 100 instances from the terminology explanation experiment and another 100 instances from the error classification experiment, where models were required to distinguish between terminology understanding errors and sociocultural inference errors. Each instance was independently evaluated by three annotators with strong familiarity with regional Chinese expressions. Annotators were allowed to consult external resources during the evaluation process to ensure the accuracy and consistency of their judgments.

For each instance, we evaluated the agreement between human annotations and several widely used commercial LLMs, including Claude-haiku-4.5, Gemini-3.1-flash, and GPT-5.4, as well as the majority-vote result across the three models. The model-human agreement results are reported in Table~\ref{tab:hm_agreement}. Our analysis shows that majority voting achieves the highest agreement with human judgments on both tasks, reaching 87\% for terminology explanation and 84\% for error classification, with Cohen’s $\kappa$ values of approximately 0.788 and 0.740, respectively. These results indicate that the multi-model majority-vote strategy provides a reliable judging mechanism and helps mitigate potential biases introduced by any single model. Therefore, we adopt three-model majority voting as the primary LLM-as-Judge mechanism for evaluating the full dataset.

\section{Supplementary Quantitative Results}
\label{sec:Supplementary Quantitative Results}

\subsection{Superficial Artifacts}
\label{sec: Superficial Artifacts}

To test whether RegioZH can be solved through superficial artifacts rather than regional language understanding, we conducted a surface-artifact control analysis, summarized in Table~\ref{tab:surface_artifact_control}.

A cross-fitted classifier using only surface and lexical features predicts the source region with 94.3\% accuracy, confirming that the items carry strong regional signal. The same features, however, do not leak the correct answer set. A surface-only answer-set classifier reaches 20.5\% exact match, identical to a baseline that predicts the region-conditional majority answer from the gold region alone, indicating that surface features contribute no answer-relevant information beyond region identity. Supplying the gold region to the surface classifier raises accuracy only to 22.1\%. A stronger diagnostic that additionally receives the gold answer cardinality reaches 31.6\%, but this baseline consumes oracle information unavailable at inference time and remains far below the best performing LLMs, which attain 71.6\% exact match.

Regional vocabularies also overlap only marginally. The mean off-diagonal Jaccard similarity is 0.004, with a maximum of 0.022 between Singapore and Malaysia. The residual region confusions concentrate on Mainland--Singapore and Singapore--Malaysia, which is linguistically expected given shared Mandarin structures and Southeast Asian lexical influence. Taken together, these results indicate that RegioZH is not reducible to region identification or lexical artifacts. Solving an item requires using regional cues to recover the meaning of localized expressions and sociocultural references, and then mapping that meaning onto the correct option set. 

\begin{table*}[htbp] 
\centering 
\begin{threeparttable} 
\begin{tabularx}{\textwidth}{ 
@{} >{\raggedright\arraybackslash}p{0.38\textwidth} >{\raggedright\arraybackslash}X >{\raggedleft\arraybackslash}p{0.18\textwidth} @{} } \toprule \textbf{Diagnostic} & \textbf{Input Information} & \textbf{Accuracy (\%)} \\ \midrule \multicolumn{3}{@{}l}{\textit{Source-region prediction}} \\ \addlinespace[2pt] Surface region classifier & Surface + lexical features & 94.3 \\ \midrule \multicolumn{3}{@{}l} {\textit{Correct answer-set prediction}} \\ \addlinespace[2pt] Uniform random exact match & None & 6.6 \\ Gold-cardinality random & Gold set size only & 21.7 \\ Global majority answer & Dataset answer prior & 12.8 \\ Region majority answer & Gold region only & 20.5 \\ Surface logistic classifier & Surface + lexical features & 20.5 \\ Surface + region classifier & Surface + lexical features + region & 22.1 \\ Surface + region + cardinality classifier & Surface + lexical features + region + gold set size & 31.6 \\ \midrule \multicolumn{3}{@{}l}{\textit{Vocabulary overlap}} \\ \addlinespace[2pt] Mean off-diagonal Jaccard & Regional vocabularies & 0.004 \\ Maximum off-diagonal Jaccard & Singapore--Malaysia & 0.02 \\ \bottomrule 
\end{tabularx} 
\end{threeparttable} 
\caption{Surface-artifact control analysis. Region prediction uses only surface and lexical features. Answer-set prediction tests whether the same features leak the correct option set.} 
\label{tab:surface_artifact_control} 
\end{table*}

\subsection{Controlling for Item Structure in Exact-Set-Match Evaluation} 
\label{sec:item-structure-controls} 
Strict exact-set-match accuracy may be affected by item structure, particularly the number of answer options \(K\) and the gold-set cardinality \(G\). We therefore examine whether cross-regional variation in these factors accounts for the observed performance disparities. Variation in \(K\) is negligible in RegioZH: the regional mean ranges from 4.00 to 4.02, and nearly all items contain exactly four options. Under uniform sampling over all non-empty answer subsets, the corresponding chance level, 

\[ c_i = \frac{1}{2^{K_i}-1}, \] 

ranges only from \(6.60\%\) to \(6.67\%\) across regions. Per-item chance normalization yields the same overall regional pattern, indicating that variation in option count does not explain the observed disparities. 

Gold-set cardinality varies more substantially across regions and may therefore affect strict exact-match accuracy. To control for this variation, we compute the stratum-specific accuracy \(A_{m,r,g}\) for each model \(m\), region \(r\), and gold-set size \(g\), and standardize all regions to the same pooled gold-cardinality distribution: 

\[ A_{m,r}^{\mathrm{std}} = \sum_g w_g A_{m,r,g}, \] 

where \(w_1=0.3509\), \(w_2=0.5626\), and \(w_3=0.0865\). The primary analysis is restricted to the common support \(G\in\{1,2,3\}\), as \(G=4\) items occur only in Hong Kong and Singapore. Standardization changes the magnitude of some regional differences: Singapore increases from \(31.6\%\) to \(39.7\%\), whereas Hong Kong decreases from \(45.3\%\) to \(35.2\%\). Nevertheless, the main regional pattern remains unchanged, with Taiwan (\(58.8\%\)), Mainland China (\(56.6\%\)), and Malaysia (\(56.2\%\)) substantially outperforming Singapore (\(39.7\%\)) and Hong Kong (\(35.2\%\)). A sensitivity analysis retaining all items and grouping the strata as \(G=1\), \(G=2\), and \(G\geq3\) produces nearly identical standardized scores. These results indicate that gold-set composition partly affects the magnitude of raw exact-match differences but does not account for the main cross-regional disparity. We therefore use exact set match as a strict primary metric while acknowledging that partial-credit metrics provide a complementary view of model performance.

\subsection{Prompt Ablations, Verbosity, and Parsing Controls} \label{sec:prompt_ablations} 
To verify that the observed elicitation effects were not artifacts of prompt construction or evaluation procedures, we conducted two complementary ablation studies across all five models and both elicitation methods. Specifically, we examined sensitivity to the ordering of answer options and to semantics-preserving variations in instruction style. We additionally controlled response verbosity and answer parsing by separating free-form reasoning from the structured answer field used for scoring. The complete ablation results are reported in Table~\ref{tab:prompt_ablation_results}, and the instruction-style paraphrases are provided in Table~\ref{tab:prompt-variants}. 

\paragraph{Option-order ablation.} We re-evaluated all ten model--method configurations after randomizing the positions of the correct option or options, while preserving the question content and evaluation procedure. As shown in the \textit{Shuffle} and $\Delta$ columns of Table~\ref{tab:prompt_ablation_results}, individual configurations changed by 1.34 accuracy points on average. The largest shift was a 3.81-point decrease for Llama-3.1-8B under Persona prompting. These changes were not systematic, as indicated by a paired $t$-test across the ten configurations, $t(9)=-0.48$, $p=0.65$, and they remained within the bootstrap confidence intervals reported for the main results. 

The option-order ablation also preserved the principal trends established by the original evaluation. The ranking of the five models remained unchanged, the performance gap between the best- and worst-performing models remained approximately 45 points, and prompting continued to outperform the corresponding zero-shot baseline in 9 of the 10 configurations. Option position therefore introduces bounded variation rather than a systematic confound. 

\paragraph{Instruction-style ablation.} To evaluate sensitivity to prompt wording, we constructed two semantics-preserving paraphrases of each original Persona and structured chain-of-thought (CoT) template. The complete paraphrased instructions are presented in Table~\ref{tab:prompt-variants}. Variant 1 and Variant 2 correspond to P2 and P3, respectively, while P1 denotes the original prompt. For Persona prompting, the variants changed the wording of the expert role and the description of regional linguistic and cultural expertise. For structured CoT prompting, the variants rephrased the region-identification, localized-term analysis, and answer-selection steps. Across all variants, we held the task semantics and final output format constant. 

As shown in the P1--P3 and $\sigma$ columns of Table~\ref{tab:prompt_ablation_results}, accuracy remained stable across the three realizations of each template. The standard deviation across prompt variants averaged 0.87 accuracy points and never exceeded 1.5 points. The relative effectiveness of the two elicitation methods was also preserved. CoT outperformed Persona for GPT-5.1, whereas Persona outperformed CoT for Ministral-3-14B and Llama-3.1-8B. For DeepSeek-V3.1 and Qwen3-30B, the differences between the two methods remained within the range of prompt-level variation. These results indicate that the observed elicitation effects cannot be attributed to a particular wording, role formulation, or step description.

\paragraph{Verbosity and parsing controls.} To ensure that differences in response length or structure did not affect evaluation, all prompt variants required the final prediction to appear in a fixed, machine-readable answer field. The answer field contained only option letters, with multiple selections represented as comma-separated lowercase letters without spaces. Explanations and intermediate analyses, when requested, were kept separate from the final answer field and were not used for scoring. Consequently, Persona and CoT responses could differ in reasoning style and verbosity without requiring method-specific answer extraction rules. 

Taken together, these ablations show that the reported results are robust to option ordering, surface-level instruction phrasing, response verbosity, and answer extraction. Option randomization introduces only bounded and unsystematic variation, while semantics-preserving prompt paraphrases leave the main model and elicitation trends unchanged. The observed effects therefore reflect stable model--method interactions rather than artifacts of a particular prompt realization or parsing convention.

\begin{table*}[t] 
\centering 
\small 
\setlength{\tabcolsep}{5.5pt} 
\begin{tabular}{llrrrrrrr} \toprule Model & Method & P1 & P2 & P3 & $\sigma$ & Shuffle & $\Delta$ & Zero-shot \\ \midrule \multirow{2}{*}{GPT-5.1} & CoT & 75.35 & 74.11 & 75.73 & 0.85 & 75.95 & $+0.60$ & 69.9 \\ & Persona & 71.88 & 68.96 & 70.12 & 1.47 & 73.95 & $+2.07$ & 69.9 \\ \addlinespace \multirow{2}{*}{DeepSeek-V3.1} & CoT & 71.97 & 74.31 & 73.66 & 1.21 & 73.16 & $+1.19$ & 71.6 \\ & Persona & 73.81 & 74.13 & 74.02 & 0.16 & 73.15 & $-0.66$ & 71.6 \\ \addlinespace \multirow{2}{*}{Qwen3-30B} & CoT & 62.67 & 64.92 & 63.58 & 1.13 & 60.32 & $-2.35$ & 61.7 \\ & Persona & 63.18 & 63.22 & 61.75 & 0.84 & 64.17 & $+0.99$ & 61.7 \\ \addlinespace \multirow{2}{*}{Ministral-3-14B} & CoT & 47.71 & 46.67 & 46.81 & 0.56 & 48.22 & $+0.52$ & 48.8 \\ & Persona & 51.42 & 52.93 & 50.78 & 1.10 & 51.06 & $-0.36$ & 48.8 \\ \addlinespace \multirow{2}{*}{Llama-3.1-8B} & CoT & 29.48 & 30.02 & 28.53 & 0.75 & 28.64 & $-0.84$ & 26.4 \\ & Persona & 33.98 & 34.13 & 33.02 & 0.60 & 30.17 & $-3.81$ & 26.4 \\ \bottomrule 
\end{tabular}
\caption{ Prompt ablation results averaged over the five regional subsets. P1 denotes the original prompt, while P2 and P3 denote two semantics-preserving paraphrases. $\sigma$ is the standard deviation across P1--P3. \textit{Shuffle} reports accuracy after randomizing the position of the correct option or options, and $\Delta$ is the difference between \textit{Shuffle} and P1. All values are accuracy points. }
\label{tab:prompt_ablation_results} 
\end{table*}

\begin{table*}[t] \centering \footnotesize \setlength{\parskip}{0pt}  \tcbset{ promptbox/.style={ colframe=black!60, boxrule=0.5pt, boxsep=1.5pt, left=3pt, right=3pt, top=2pt, bottom=2pt, before skip=1pt, after skip=1pt, toptitle=2pt, bottomtitle=2pt, fonttitle=\bfseries, breakable } } \begin{tcolorbox}[ promptbox, colback=gray!5, title=Persona Prompting: Variant 1 (P2) ] Act as an expert in the Chinese language and culture of the \texttt{\{region\}} region. You are highly familiar with region-specific expressions, colloquial usage, cultural conventions, and the intended meanings of everyday communication.\\[2pt] Read the following question carefully and identify all correct option(s). More than one option may be correct. Briefly explain why you selected those option(s).\\[2pt] \texttt{\{question\_and\_answer\}}\\[2pt] Use the following response format:\\[1pt] \texttt{ANSWER: [letter(s) of correct option(s), e.g., ``a'' or ``a,c'']} \end{tcolorbox} \vspace{1pt} \begin{tcolorbox}[ promptbox, colback=gray!5, title=Persona Prompting: Variant 2 (P3) ] You have specialized expertise in the linguistic and cultural practices of the \texttt{\{region\}} region, including local varieties of Chinese, informal expressions, culturally specific knowledge, and pragmatic interpretation in daily conversations.\\[2pt] Based on this expertise, answer the question below by choosing the correct option or options and explaining the reasoning behind your choice. The question may have multiple correct answers.\\[2pt] \texttt{\{question\_and\_answer\}}\\[2pt] Return your answer in this format:\\[1pt] \texttt{ANSWER: [letter(s) of correct option(s), e.g., ``a'' or ``a,c'']} \end{tcolorbox} \vspace{1pt} \begin{tcolorbox}[ promptbox, colback=yellow!5, title=Structured CoT Prompting: Variant 1 (P2) ] You are analyzing a Chinese multiple-choice question that may contain region-specific linguistic features.\\[2pt] Evaluate the question by following these steps: \begin{enumerate}[ leftmargin=*, itemsep=0pt, parsep=0pt, topsep=1pt, partopsep=0pt ] \item \textbf{Region:} Determine which region the question most likely originates from: Mainland China, Hong Kong, Taiwan, Singapore, or Malaysia. \item \textbf{Localized Terms:} Identify any localized expressions, such as dialectal words, slang, or transliterations. For each expression, provide its standard Chinese equivalent or a brief explanation. If there are no localized terms, output \texttt{- None}. \item \textbf{Answer:} Choose all correct options according to the linguistic meaning of the question. Do not rely on regional stereotypes. \end{enumerate} \texttt{\{question\_and\_answer\}}\\[2pt] Follow the exact output format below. The final Answer must contain only lowercase option letters. If multiple options are correct, separate them with commas and do not include spaces, e.g., \texttt{\{\{a,c\}\}}.\\[2pt] \texttt{Region:}\\ \texttt{Localized Terms:}\\ \texttt{\textbullet\ - $\rightarrow$}\\ \texttt{Answer: \{\{\}\}} \end{tcolorbox} \vspace{1pt} \begin{tcolorbox}[ promptbox, colback=yellow!5, title=Structured CoT Prompting: Variant 2 (P3) ] Your task is to solve a Chinese multiple-choice question by first examining its regional linguistic characteristics.\\[2pt] Process the question in the following order: \begin{enumerate}[ leftmargin=*, itemsep=0pt, parsep=0pt, topsep=1pt, partopsep=0pt ] \item \textbf{Region:} Infer the most probable regional source from the following candidates: Mainland China, Hong Kong, Taiwan, Singapore, or Malaysia. \item \textbf{Localized Terms:} Find expressions that are specific to a particular region, including dialect forms, colloquial slang, and transliterated words. Convert each expression into standard Chinese or briefly describe its meaning. If no such expressions are present, output \texttt{- None}. \item \textbf{Answer:} Select every correct option based only on the meaning conveyed by the language, without making assumptions based on regional stereotypes. \end{enumerate} \texttt{\{question\_and\_answer\}}\\[2pt] Return the result strictly in the template below. In the final Answer field, use lowercase letters only. For multiple correct options, use comma-separated letters with no spaces, for example \texttt{\{\{a,c\}\}}.\\[2pt] \texttt{Region:}\\ \texttt{Localized Terms:}\\ \texttt{\textbullet\ - $\rightarrow$}\\ \texttt{Answer: \{\{\}\}} \end{tcolorbox} \caption{ Semantics-preserving prompt variants used in the instruction-style ablation. Variant 1 and Variant 2 correspond to P2 and P3, respectively, in Table~\ref{tab:prompt_ablation_results}, while P1 denotes the original prompt. The field \texttt{\{question\_and\_answer\}} is filled with the multiple-choice question and its answer options at inference time. The Persona variants additionally use the target-region field \texttt{\{region\}}. Surface wording, role framing, and step descriptions vary across prompt variants, while the task semantics and final answer format remain fixed. } \label{tab:prompt-variants} \end{table*}

\subsection{Cross-Model Diagnostic Replication}
\label{sec: Results of Qwen and SeaLLMs}

\paragraph{Representation Geometry and Regional Separation.}

To examine whether the representation-level pattern observed in Section~\ref{sec:Representation Geometry} generalizes beyond GLM-4, we repeat the semantic-space analysis on Qwen3 and SeaLLMs-V3. We construct a paired representation space that directly compares region-specific inputs against their standardized counterparts. For each evaluated model, we formulate a consolidated text corpus by aggregating informal instances and their formal equivalents across all five target regions: Mainland China, Singapore, Taiwan, Hong Kong, and Malaysia. To obtain the formal equivalents, we used the prompt provided in Table~\ref{tab:rewrite_prompt} to instruct Claude-haiku-4.5 to convert informal region-specific Chinese expressions into standardized Chinese expressions. The converted outputs were then reviewed and revised by annotators from the corresponding regions, who conducted external knowledge searches when necessary to reduce interpretive uncertainty and ensure the fluency, accuracy, and semantic consistency of the standardized texts. Formally, we define this dataset as:
\begin{equation}
\mathcal{D}={(x_i, r_i, s_i)}{i=1}^{N}
\end{equation}
where $x_i$ represents the input text, $r_i$ denotes the region label, and $s_i \in \{\text{informal}, \text{formal}\}$ indicates the stylistic formulation. To construct the semantic space, we extract sentence-level embeddings from the final hidden layer of each respective model. Given a tokenized input sequence $x_i \rightarrow (t{i1}, t_{i2}, \ldots, t_{iL_i})$, we perform a forward pass to obtain the hidden states $H_i=[h_{i1}, h_{i2}, \ldots, h_{iL_i}] \in \mathbb{R}^{L_i \times d}$, where $d$ is the model's hidden dimension. To derive a single vector representation while ignoring padded tokens, we apply attention-mask-aware mean pooling over the sequence:
\begin{equation}
\bar{h}i = \frac{\sum{j=1}^{L_i} m_{ij} h_{ij}}{\sum_{j=1}^{L_i} m_{ij}}
\end{equation}
where $m_{ij} \in \{0,1\}$ acts as the attention mask corresponding to the $j$-th token. This pooled representation is subsequently $L_2$-normalized to yield the final semantic vector:
\begin{equation}
e_i = \frac{\bar{h}_i}{||\bar{h}_i||_2}
\end{equation}
These vectors are compiled across the entire dataset to produce the global embedding matrix $E \in \mathbb{R}^{N \times d}$.  To visualize and quantify regional semantic separation, we project these high-dimensional embeddings into a shared three-dimensional space. Crucially, rather than fitting separate transformations for each region, we fit a single Principal Component Analysis (PCA) jointly over the concatenated embeddings of all regions and styles within a given model, ensuring that the resulting geometric distances are directly comparable. We centralize the global embedding matrix as $\tilde{E}=E-\mathbf{1}\mu^\top$, where $\mu=\frac{1}{N}\sum_{i=1}^{N} e_i$, and compute the covariance matrix:
\begin{equation}
C = \frac{1}{N-1}\tilde{E}^{\top}\tilde{E}
\end{equation}
By solving for the eigenvectors corresponding to the top three eigenvalues ($\lambda_1 \geq \lambda_2 \geq \lambda_3$), we obtain the projection matrix $W_3=[v_1, v_2, v_3] \in \mathbb{R}^{d \times 3}$. The projected 3D coordinates for each sample are therefore given by:
\begin{equation}
z_i = \tilde{e}i W_3 = (PC1_i, PC2_i, PC3_i)
\end{equation}
where the first principal component (PC1) captures the maximum global semantic variance. Finally, to systematically analyze the geometric displacement between informal regional expressions and their standardized equivalents, we compute region-style centroids by averaging the projected coordinates:
\begin{equation}
c{r,s} = \frac{1}{|\mathcal{I}{r,s}|} \sum{i \in \mathcal{I}{r,s}} z_i
\end{equation}
where $\mathcal{I}{r,s}={i \mid r_i=r, s_i=s}$. By mapping these centroids and their respective data manifolds within the same projected space, we can observe whether lexical normalization successfully collapses geometric distance, or if structural and contact-induced written features cause certain regions to remain uniquely isolated.

\begin{figure}[htpb]
  \centering
  \captionsetup{skip=0pt}
  \includegraphics[trim=50 40 50 20, width=0.45\textwidth]{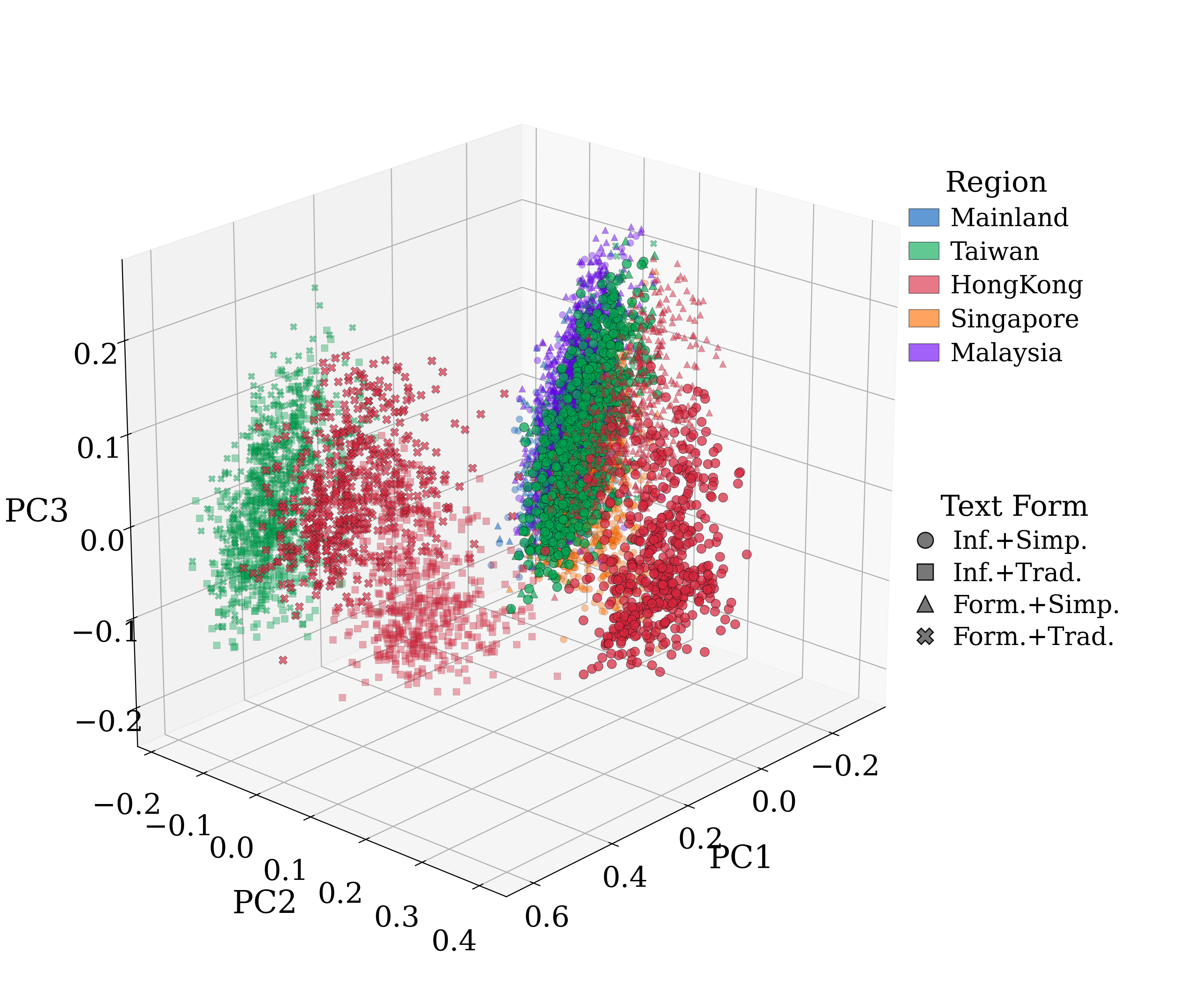}
  \caption{Semantic space visualization. PCA projection of regional expressions and their standardized counterparts (Qwen).}
  \label{fig:semantic_representation_qwen}
\end{figure}

\begin{figure}[htpb]
  \centering
  \captionsetup{skip=0pt}
  \includegraphics[trim=50 40 50 20, width=0.45\textwidth]{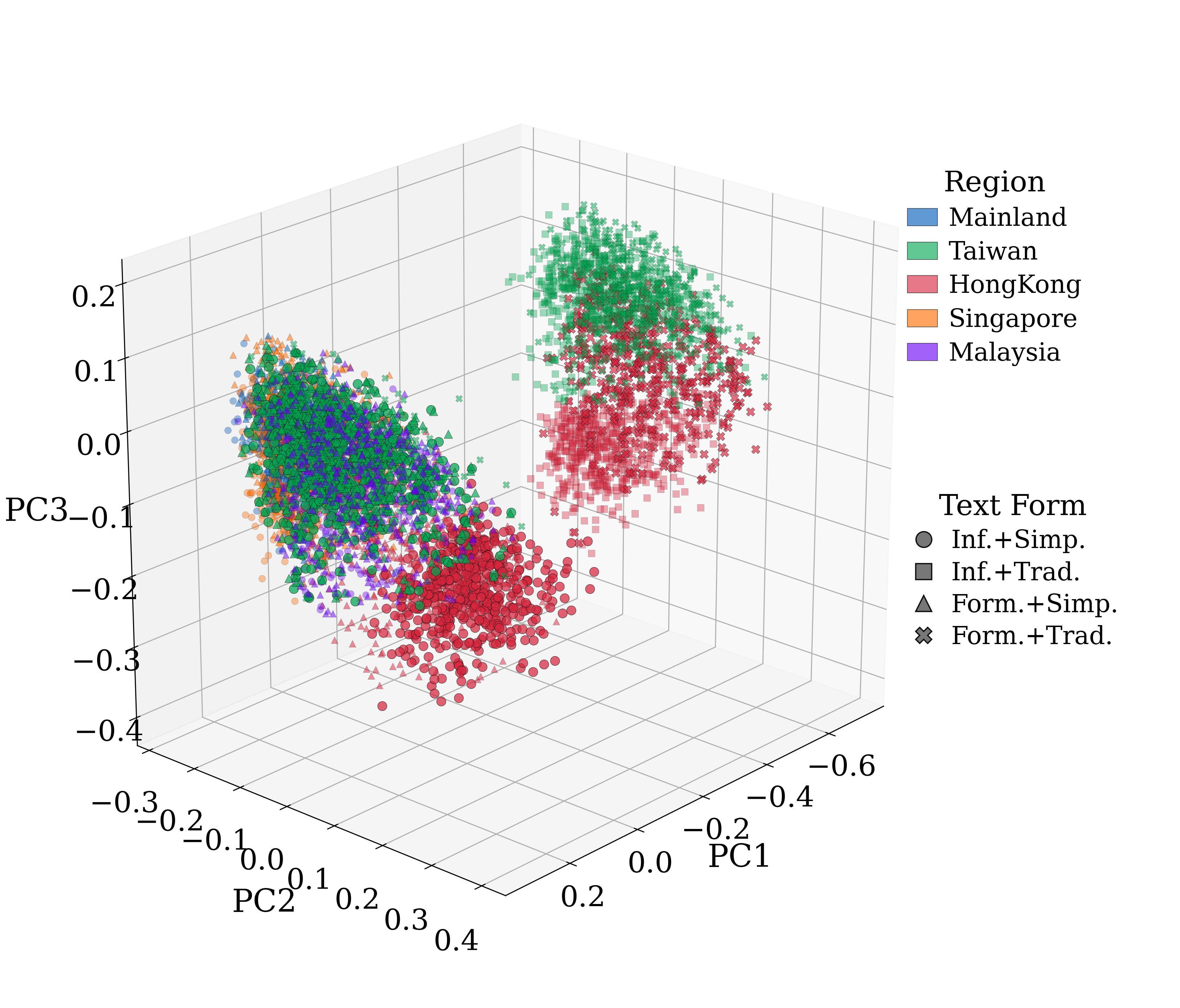}
  \caption{Semantic space visualization. PCA projection of regional expressions and their standardized counterparts (SeaLLM).}
  \label{fig:semantic_representation_seallm}
\end{figure}

As shown in Figures~\ref{fig:semantic_representation_qwen} and~\ref{fig:semantic_representation_seallm}, Qwen3 and SeaLLMs-V3 exhibit representation-geometry patterns that closely mirror the GLM-4 results in Section~\ref{sec:Representation Geometry}. Under the unified Simplified-script setting, formalized expressions from all regions occupy relatively proximate semantic spaces. For Qwen3, the formal MMD$^2$ values range from 0.038 to 0.114 (Tables~\ref{tab:mmd_qwen}), and for SeaLLMs-V3 they range from 0.039 to 0.106 (Tables~\ref{tab:mmd_seallm}). The same convergence is also observed for most informal Simplified inputs: Taiwan, Singapore, Mainland China, and Malaysia remain below 0.15 in both models. Hong Kong is the clear exception. Its informal Simplified MMD$^2$ reaches 0.602 for Qwen3 and 0.693 for SeaLLMs-V3, far above the corresponding values for other regions. This reproduces the GLM-4 pattern, where Hong Kong informal expressions remain geometrically separated even after conversion to Simplified Chinese.

The Traditional-script and cross-script conditions further clarify this effect. For Hong Kong, the original informal Traditional inputs remain strongly separated in both Qwen3 and SeaLLMs-V3, with MMD$^2$ values of 1.156 and 1.267, respectively. In contrast, formalizing Hong Kong inputs into a more standardized written form substantially reduces their distance when the output remains in Traditional script, yielding values of 0.237 for Qwen3 and 0.220 for SeaLLMs-V3. Taiwan shows a different pattern: although its Traditional inputs can be separated at the representation level, its formalized Traditional condition is much closer to the corresponding informal Traditional space, especially under the cross-condition comparison, where both Qwen3 and SeaLLMs-V3 obtain an MMD$^2$ value of 0.097. This contrast supports the claim in Section~\ref{sec:Representation Geometry} that Hong Kong and Taiwan Traditional Chinese should not be treated as the same surface phenomenon.

Overall, the additional Qwen3 and SeaLLMs-V3 results strengthen the interpretation drawn from GLM-4. Script variation is a salient organizing factor in the embedding space, but it does not fully explain regional difficulty. Formalization and script unification generally collapse regional distances, suggesting that the models encode standardized Chinese varieties more similarly. However, Hong Kong remains an outlier across all three representative models, indicating that its difficulty is tied not only to Traditional script but also to Cantonese-influenced vocabulary, syntax, and Hong Kong-specific written conventions. By contrast, the smaller MMD$^2$ values for Taiwan, Singapore, Mainland China, and Malaysia suggest that their errors are less likely to arise from global representation displacement alone, and are more plausibly associated with localized lexical grounding, sociocultural inference, and downstream answer selection.

\begin{table*}[t]
\centering
\small
\begin{tcolorbox}[colback=blue!3, colframe=black!60,
  title=\textbf{User Prompt Template (shared across all regions)}]
Please output the rewritten full text directly.\\[4pt]
Text to rewrite:\\
\{\{question\_and\_answer\}\}\\[4pt]
Reference correct answer (for preserving question logic only; do not mention it in the output):\\
\{\{reference\_answer\}\} \quad \textit{(optional; included only when available)}
\end{tcolorbox}
\begin{tcolorbox}[colback=gray!5, colframe=black!60,
  title=\textbf{System Prompt (region-specific variants)}]
You are a Chinese editor and localization rewriting assistant.
Please rewrite the input text into formal, standard, and natural modern written Chinese
(Simplified Chinese).\\[6pt]
\textbf{Requirements:}
\begin{enumerate}\itemsep2pt
  \item \textit{(Region-specific — see region table below)}
        Remove or rewrite the region-specific informal expressions and
        replace them with semantically equivalent formal written Chinese, choosing the most
        natural and context-appropriate wording.
  \item Preserve the original meaning, question type, forms of address,
        question-and-answer logic, number of options, option labels (such as
        \texttt{(a)(b)(c)}), and the intended direction of correct/incorrect understanding.
  \item Do not add explanations, analysis, background information, or remove any options.
  \item If the original text is already formal, make only the minimum necessary revisions.
  \item Maintain a neutral, clear, and polished written style suitable for educational,
        professional, or formal written contexts.
  \item Do not use archaic, literary, overly bureaucratic, or internet-style language.
  \item Output only the fully rewritten text, without titles, quotation marks, notes, or
        any extra content.
\end{enumerate}
\textit{Note: For Hong Kong data, an additional requirement is prepended: ``Convert
Traditional Chinese characters to Simplified Chinese.'' The Hong Kong variant also adds a
requirement to preserve proper nouns, exam names, school system terms, or institutional
terms, rewriting them into the most understandable standard written Chinese form only when
necessary.}
\end{tcolorbox}
\begin{tcolorbox}[colback=yellow!5, colframe=black!60,
  title=\textbf{Region-Specific Expression Types for Requirement 1}]
\begin{tabular}{lp{0.75\textwidth}}
\toprule
\textbf{Region} & \textbf{Expression types to remove / rewrite} \\
\midrule
Mainland China  & Colloquial Mainland expressions, internet slang, trendy online
                  phrases, chat-style abbreviations, filler words, spoken discourse
                  markers, exaggerated emotional wording \\[4pt]
Singapore       & Singapore-style expressions, Singlish, mixed English, Hokkien,
                  Fujian dialect, Malay, and other localized expressions \\[4pt]
Taiwan          & Taiwan-style colloquial expressions, informal spoken Mandarin,
                  internet slang, mixed Hokkien/Southern~Min expressions, chat-style
                  abbreviations, filler words \\[4pt]
Malaysia        & Malaysian-style expressions, mixed English, Malay loanwords,
                  Cantonese- or Hokkien-influenced expressions, transliterated local
                  terms, sentence-final particles \\[4pt]
Hong Kong       & Hong Kong-style expressions, Cantonese colloquialisms, Hong
                  Kong-specific spoken vocabulary, sentence-final particles, mixed
                  English, transliterated Cantonese expressions \\
\bottomrule
\end{tabular}
\end{tcolorbox}
\caption{Prompt template used to instruct Claude-haiku-4.5 to convert region-specific informal
Chinese expressions into standardized Chinese. The system prompt varies by region only in
the first requirement (italicized); all other requirements are shared. The user prompt
template is identical across all five regions. \texttt{\{\{...\}\}} denotes fields filled
at inference time; the reference answer field is omitted when unavailable.}
\label{tab:rewrite_prompt}
\end{table*}

\begin{table}[t]
\centering
\small
\setlength{\tabcolsep}{3pt}
\renewcommand{\arraystretch}{1.05}

\begin{tabular}{l ccccc cc}
\toprule
& \multicolumn{5}{c}{Simplified} & \multicolumn{2}{c}{Traditional} \\
\cmidrule(lr){2-6} \cmidrule(lr){7-8}
Register & HK & TW & SG & ML & MY & HK & TW \\
\midrule
Informal & 0.602 & 0.070 & 0.085 & 0.144 & 0.107 & 1.156 & 0.940 \\
Formal   & 0.114 & 0.038 & 0.060 & 0.066 & 0.068 & 1.075 & 0.622 \\
\bottomrule
\end{tabular}
\caption{Qwen cross-regional MMD$^2$ distances. Each cell reports the unbiased MMD$^2$ between the focal region (column) and the pooled distribution of the remaining four regions, under the matching register. The reference distribution is always in Simplified; the right-hand panel keeps HK and TW in Traditional. Lower values indicate greater proximity to the rest of the corpus.}
\label{tab:mmd_qwen}
\end{table}

\begin{table}[t]
\centering
\small
\setlength{\tabcolsep}{3pt}
\renewcommand{\arraystretch}{1.05}

\begin{tabular}{l ccccc cc}
\toprule
& \multicolumn{5}{c}{Simplified} & \multicolumn{2}{c}{Traditional} \\
\cmidrule(lr){2-6} \cmidrule(lr){7-8}
Register & HK & TW & SG & ML & MY & HK & TW \\
\midrule
Informal & 0.693 & 0.071 & 0.099 & 0.140 & 0.096 & 1.267 & 0.966 \\
Formal   & 0.106 & 0.039 & 0.070 & 0.063 & 0.049 & 1.266 & 0.697 \\
\bottomrule
\end{tabular}
\caption{SeaLLM cross-regional MMD$^2$ distances. Each cell reports the unbiased MMD$^2$ between the focal region (column) and the pooled distribution of the remaining four regions, under the matching register. The reference distribution is always in Simplified; the right-hand panel keeps HK and TW in Traditional. Lower values indicate greater proximity to the rest of the corpus.}
\label{tab:mmd_seallm}
\end{table}

\begin{figure}[htpb]
  \centering
  \captionsetup{skip=0pt}
  \includegraphics[width=0.45\textwidth]{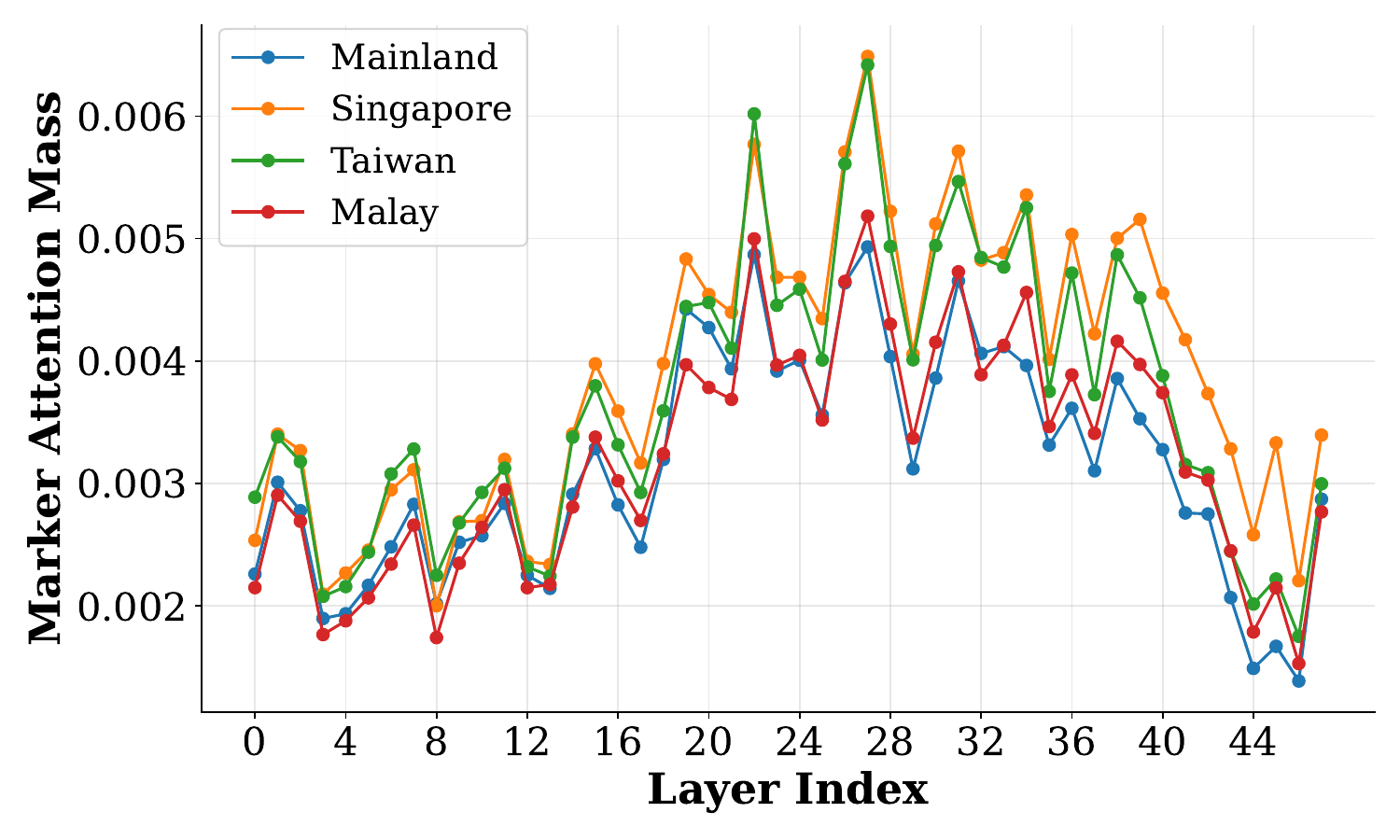}
  \caption{Layer-wise attention mass assigned to region-specific marker tokens across regional Chinese varieties (Qwen).}
  \label{fig:Model attention by layers Qwen}
\end{figure}

\begin{figure}[htpb]
  \centering
  \captionsetup{skip=0pt}
  \includegraphics[width=0.45\textwidth]{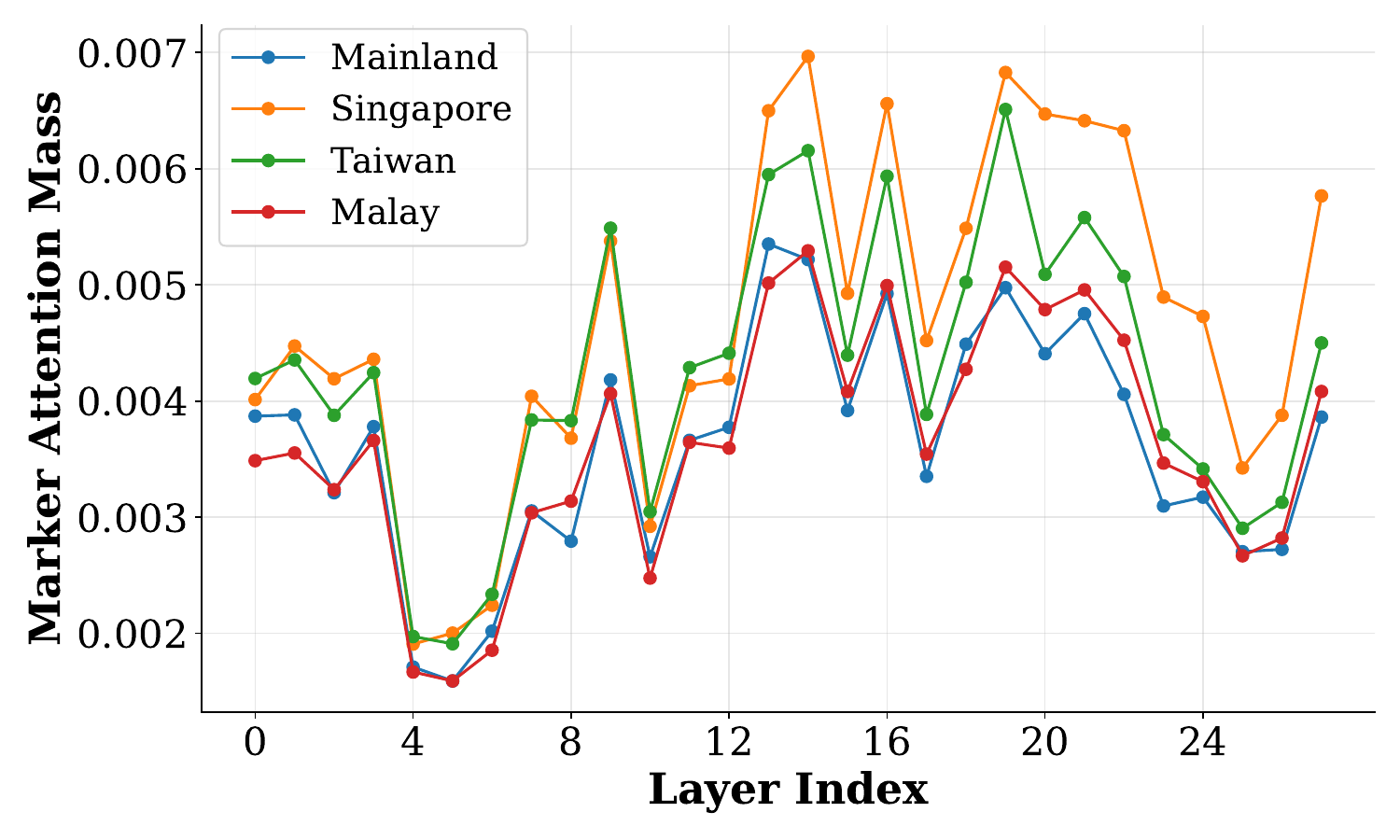}
  \caption{Layer-wise attention mass assigned to region-specific marker tokens across regional Chinese varieties (SeaLLM).}
  \label{fig:Model attention by layers seallm}
\end{figure}

\paragraph{Attention–Performance Mismatch.} To assess whether the attention-based finding in Section~\ref{sec:Attention without Grounding} generalizes beyond GLM-4, we repeat the marker-level attention analysis on Qwen3 and SeaLLMs-V3. As shown in Figures~\ref{fig:Model attention by layers Qwen} and~\ref{fig:Model attention by layers seallm}, both models assign non-negligible attention mass to region-specific marker tokens, especially in middle-to-late layers. This indicates that localized lexical cues are not simply ignored during model processing.

However, Tables~\ref{tab:regional_marker_stats_1} and~\ref{tab:regional_marker_stats_2} show that marker salience does not reliably translate into semantic grounding. For Qwen3, Singapore receives the highest average marker attention among the four analyzed regions (0.0034), but it has the lowest gold-answer probability (0.178), the lowest explanation accuracy on incorrect cases ($E_{\times}=33.0$), and the highest grounding-failure share ($G=0.67$). By contrast, Taiwan receives slightly lower marker attention (0.0030), yet achieves the highest gold-answer probability (0.397), the highest $E_{\times}$ (57.2), and the lowest grounding-failure share ($G=0.43$). SeaLLMs-V3 shows the same pattern more strongly: Singapore again has the highest marker attention (0.0058), but the lowest gold-answer probability (0.058), very low $E_{\times}$ (8.1), and the highest grounding-failure share ($G=0.92$), whereas Taiwan has lower marker attention (0.0045) but substantially better gold-answer probability (0.291) and lower grounding-failure share ($G=0.79$).

The Mainland subset further suggests that dense localized cues can introduce an additional compositional burden. Mainland items contain the highest marker density in both tables (2.92 terms per sample). For Qwen3, the model often explains terms correctly on correct items ($E_{\checkmark}=85.8$), but its gold-answer probability remains lower than Taiwan's, suggesting that recognizing individual markers is not always sufficient for composing the final region-licensed interpretation. For SeaLLMs-V3, both explanation accuracy and gold-answer probability are much lower, indicating a stronger breakdown under the same dense-marker condition.

The Qwen3 and SeaLLMs-V3 results corroborate the conclusion of Section~\ref{sec:Attention without Grounding} rather than introducing a separate pattern. Across model capability levels, regional Chinese failures are not primarily explained by an attention blind spot. Instead, models often detect localized markers but fail to consistently ground them in region-appropriate meanings and propagate those meanings into downstream answer selection.

\begin{table}[htpb]
\centering
\small
\setlength{\tabcolsep}{3.5pt}
\begin{tabular}{lrrrrrrr}
\toprule
Region & Attn. & GoldP & T/S
  & $E_{\checkmark}$ & $E_{\times}$ & $\Delta$ & $G$ \\
\midrule
Mainland  & 0.0029 & 0.365 & 2.92 & 85.8 & 48.8 & 37.1 & 0.51 \\
Singapore & 0.0034 & 0.178 & 1.62 & 59.6 & 33.0 & 26.5 & 0.67 \\
Taiwan    & 0.0030 & 0.397 & 1.64 & 82.2 & 57.2 & 25.0 & 0.43 \\
Malaysia  & 0.0028 & 0.340 & 1.51 & 69.4 & 42.1 & 27.4 & 0.58 \\
\bottomrule
\end{tabular}
\caption{Marker attention, gold-answer probability, density, and
term-explanation accuracy conditioned on answer correctness (Qwen3-30B).
$E_{\checkmark}$/$E_{\times}$: explanation accuracy
on correct/incorrect items; $\Delta = E_{\checkmark} - E_{\times}$
(pp); $G$: grounding-failure share among wrong-answer items,
$d/(c+d)$.}
\label{tab:regional_marker_stats_1}
\end{table}


\begin{table}[htpb]
\centering
\small
\setlength{\tabcolsep}{3.5pt}
\begin{tabular}{lrrrrrrr}
\toprule
Region & Attn. & GoldP & T/S
  & $E_{\checkmark}$ & $E_{\times}$ & $\Delta$ & $G$ \\
\midrule
Mainland  & 0.0039 & 0.174 & 2.92 & 31.7 & 12.7 & 18.9 & 0.87 \\
Singapore & 0.0058 & 0.058 & 1.62 & 10.3 &  8.1 &  2.2 & 0.92 \\
Taiwan    & 0.0045 & 0.291 & 1.64 & 46.2 & 20.6 & 25.6 & 0.79 \\
Malaysia  & 0.0041 & 0.254 & 1.51 & 36.1 & 16.6 & 19.5 & 0.83 \\
\bottomrule
\end{tabular}
\caption{Marker attention, gold-answer probability, density, and
term-explanation accuracy conditioned on answer correctness (SeaLLMs-v3-7B).
$E_{\checkmark}$/$E_{\times}$: explanation accuracy
on correct/incorrect items; $\Delta = E_{\checkmark} - E_{\times}$
(pp); $G$: grounding-failure share among wrong-answer items,
$d/(c+d)$.}
\label{tab:regional_marker_stats_2}
\end{table}

\paragraph{Forced-Choice Calibration across Regions.}

We further extend the forced-choice decoding calibration analysis to Qwen3 and SeaLLMs-V3 to examine whether the calibration pattern identified in Section~\ref{sec:Confidence without Understanding} generalizes across model families. For each multiple-choice question, we evaluate the set of all valid answer combinations, denoted as $\mathcal{Y}$, including single options, option pairs, higher-order combinations, and all-choice candidates. 
Given a prompt $x$, we compute the raw conditional log-likelihood $s(y)$ for each candidate sequence $y \in \mathcal{Y}$. 
These candidate scores are then normalized into predictive probabilities using a temperature-scaled softmax:

\begin{equation}
p(y \mid x; T) = 
\frac{\exp(s(y) / T)}
{\sum_{y' \in \mathcal{Y}} \exp(s(y') / T)} ,
\end{equation}

where $T$ denotes the temperature parameter. We set $T=1.0$, which corresponds to the original forced-choice confidence, while varying $T$ provides a post-hoc transformation of the candidate score distribution. We restrict the temperature range to $T \in [0.5, 2.0]$ to avoid extremely low temperatures. Such temperatures would artificially sharpen the distribution, amplify score gaps, and push the top-candidate probability toward one, causing the scaling operation itself to dominate the Expected Calibration Error (ECE)~\citep{guo2017calibration} rather than revealing the model's native decoding confidence. Predictions correspond to the highest-probability candidate, with accuracy measured by exact match against the gold standard. Because this protocol ranks candidate answer combinations by conditional likelihood rather than parsing generated responses, the resulting forced-choice accuracy is used only to contextualize calibration metrics and is not directly comparable to the generated-answer zero-shot accuracy reported in Tab.~\ref{tab:zero-shot-result}.

As shown in Figures 14 and 15, both models exhibit overconfident miscalibration. Increasing the temperature $T$ generally softens the predictive distributions and reduces calibration errors on average, while leaving exact-match accuracy unchanged. However, the reductions are not strictly monotonic for every SeaLLM region-metric pair, especially at higher temperatures. This indicates that temperature scaling primarily softens the models' predictive distributions rather than improving their underlying regional comprehension.

As shown in Table~\ref{tab:decoding_calibration_qwen}, for Qwen3, the Singapore subset constitutes the most pronounced case of confidence misalignment. At $T=1.0$, Qwen3 achieves only 6.41\% accuracy on Singapore items, yet assigns an average confidence of 0.7894, resulting in the highest ECE, multiclass Brier score, and NLL among all regions, at 0.7253, 1.4368, and 5.9073, respectively. Although increasing the temperature to $T=2.0$ reduces the severity of miscalibration, the ECE remains substantially elevated at 0.5935, suggesting that the model continues to place excessive probability mass on incorrect candidates. A similar pattern emerges for SeaLLMs-V3, as reported in Table~\ref{tab:decoding_calibration_seallm}, albeit with a weaker degree of overconfidence. Its average ECE decreases from 0.2519 at $T=1.0$ to 0.1034 at $T=2.0$, showing that the forced-choice distribution becomes better calibrated after temperature scaling. Nevertheless, Singapore again remains the most difficult region: at $T=1.0$, SeaLLMs-V3 obtains an ECE of 0.4240, a Brier score of 1.1123, and an NLL of 3.0376, despite an accuracy of only 3.50\% and an average gold-answer probability of 0.0928.

These results reinforce the conclusion of Section~\ref{sec:Confidence without Understanding}, indicating that regional comprehension failures are accompanied by confidence distortions that are not specific to GLM-4, but recur across models with different capability levels. Temperature scaling can partially correct the probabilistic sharpness of the output distribution, yet it cannot alter the ranking of candidate answers or repair upstream failures in regional identification, localized lexical interpretation, and compositional reasoning. The calibration gap therefore provides a cross-model probabilistic signature of confidence without understanding in regional Chinese reasoning. We provide the exact numerical values corresponding to Figure~\ref{fig:all} for GLM4 in Section~\ref{sec:Confidence without Understanding} in Table~\ref{tab:decoding_calibration_glm}.

\begin{figure}[htpb]
  \centering
  \captionsetup{skip=0pt}
  \includegraphics[width=0.45\textwidth]{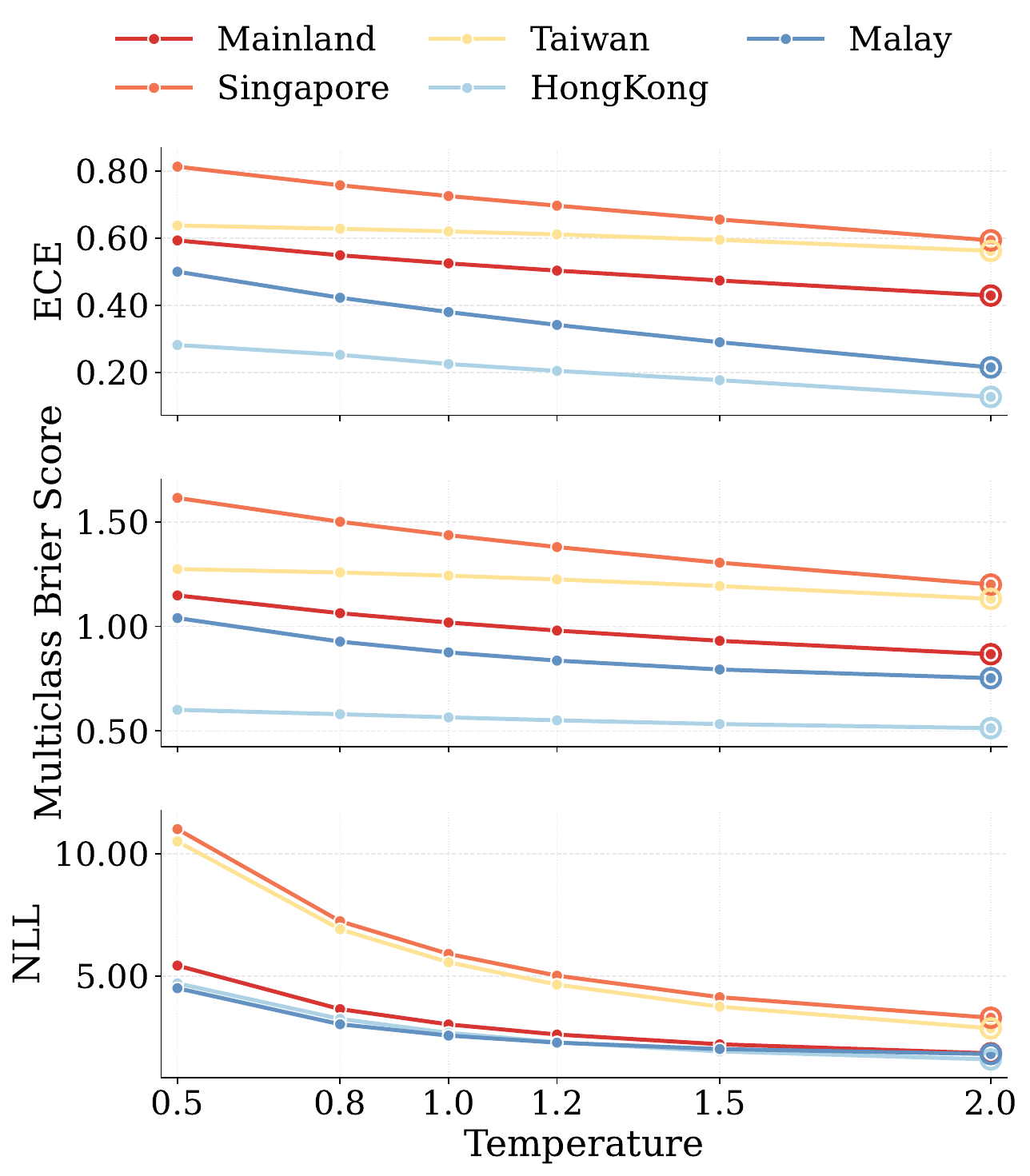}
  \caption{Temperature-scaled decoding calibration using forced-choice candidate likelihoods (Qwen).}
  \label{fig:all qwen}
\end{figure}

\begin{figure}[htpb]
  \centering
  \captionsetup{skip=0pt}
  \includegraphics[width=0.45\textwidth]{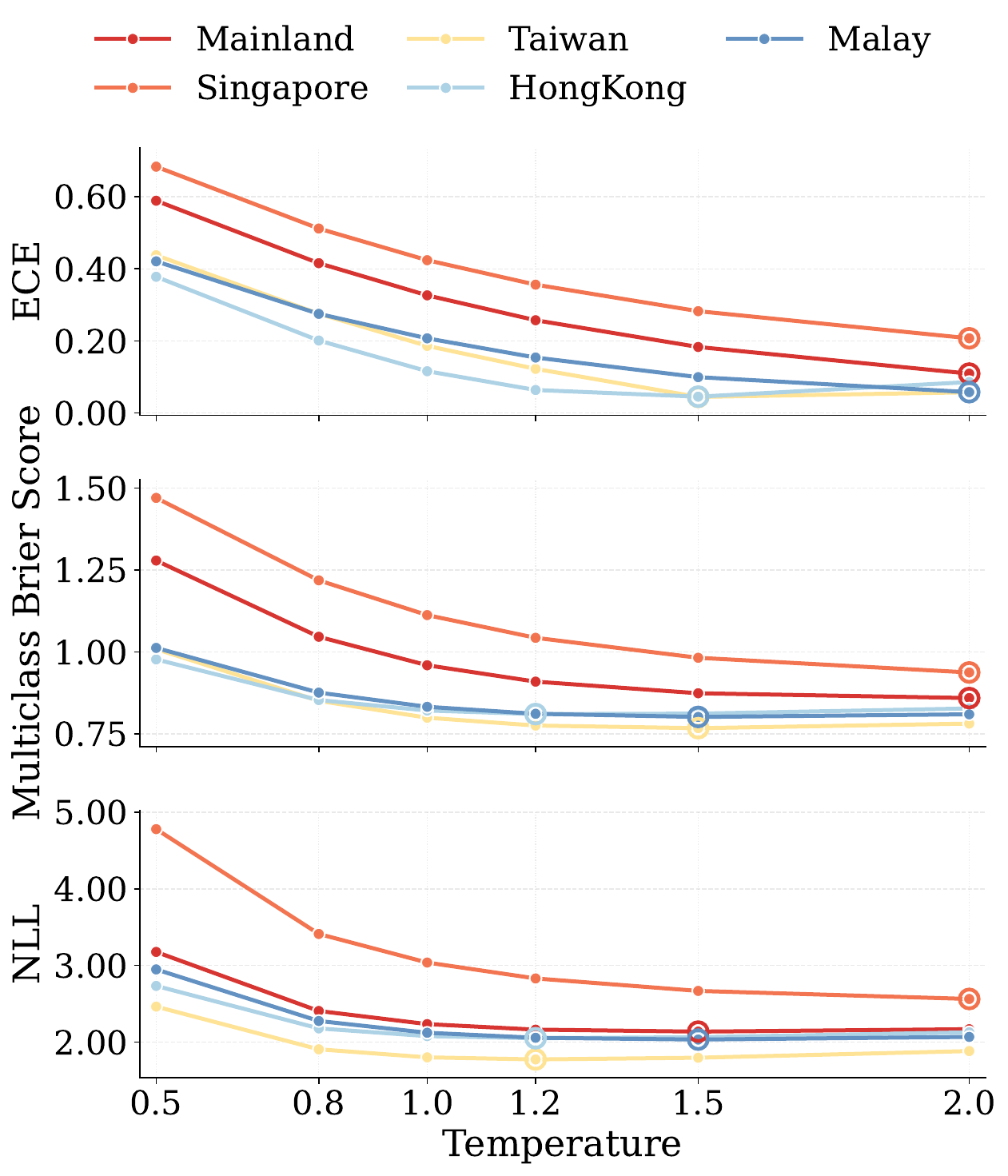}
  \caption{Temperature-scaled decoding calibration using forced-choice candidate likelihoods (SeaLLM).}
  \label{fig:all seallm}
\end{figure}

\begin{table*}[t]
\centering
\small
\setlength{\tabcolsep}{5pt}
\renewcommand{\arraystretch}{1.08}
\resizebox{\textwidth}{!}{%
\begin{tabular}{lrrrrr}
\toprule
\textbf{Region} & \textbf{Samples} & \textbf{Temperature} & \textbf{ECE} & \textbf{Multiclass Brier Score} & \textbf{Negative Log Likelihood} \\
\midrule
\multirow{6}{*}{Mainland} & 520 & 0.5 & 0.6567 & 1.2229 & 3.3926 \\
 & 520 & 0.8 & 0.5683 & 1.0454 & 2.3376 \\
 & 520 & 1.0 & 0.5181 & 0.9644 & 2.0218 \\
 & 520 & 1.2 & 0.4717 & 0.9025 & 1.8355 \\
 & 520 & 1.5 & 0.4071 & 0.8354 & 1.6864 \\
 & 520 & 2.0 & 0.3131 & 0.7727 & 1.6096 \\
\addlinespace[2pt]
\multirow{6}{*}{Singapore} & 515 & 0.5 & 0.8077 & 1.6198 & 7.6123 \\
 & 515 & 0.8 & 0.7150 & 1.4476 & 4.9592 \\
 & 515 & 1.0 & 0.6577 & 1.3510 & 4.1185 \\
 & 515 & 1.2 & 0.6036 & 1.2667 & 3.5871 \\
 & 515 & 1.5 & 0.5286 & 1.1617 & 3.0986 \\
 & 515 & 2.0 & 0.4227 & 1.0386 & 2.6886 \\
\addlinespace[2pt]
\multirow{6}{*}{Taiwan} & 536 & 0.5 & 0.5322 & 1.1150 & 3.8330 \\
 & 536 & 0.8 & 0.4462 & 0.9851 & 2.5515 \\
 & 536 & 1.0 & 0.3884 & 0.9114 & 2.1669 \\
 & 536 & 1.2 & 0.3330 & 0.8524 & 1.9381 \\
 & 536 & 1.5 & 0.2566 & 0.7896 & 1.7490 \\
 & 536 & 2.0 & 0.1754 & 0.7359 & 1.6325 \\
\addlinespace[2pt]
\multirow{6}{*}{Hong Kong} & 520 & 0.5 & 0.3744 & 0.9124 & 3.8896 \\
 & 520 & 0.8 & 0.2680 & 0.8212 & 2.6921 \\
 & 520 & 1.0 & 0.2066 & 0.7829 & 2.3412 \\
 & 520 & 1.2 & 0.1520 & 0.7578 & 2.1374 \\
 & 520 & 1.5 & 0.0916 & 0.7378 & 1.9752 \\
 & 520 & 2.0 & 0.0703 & 0.7333 & 1.8850 \\
\addlinespace[2pt]
\multirow{6}{*}{Malaysia} & 516 & 0.5 & 0.5685 & 1.2749 & 6.1179 \\
 & 516 & 0.8 & 0.4639 & 1.1334 & 4.1118 \\
 & 516 & 1.0 & 0.4064 & 1.0648 & 3.4880 \\
 & 516 & 1.2 & 0.3564 & 1.0118 & 3.0990 \\
 & 516 & 1.5 & 0.2924 & 0.9546 & 2.7466 \\
 & 516 & 2.0 & 0.2245 & 0.8994 & 2.4584 \\
\bottomrule
\end{tabular}%
}
\caption{Exact Numerical Values for Forced Choice Decoding Calibration on GLM.}
\label{tab:decoding_calibration_glm}
\end{table*}

\begin{table*}[t]
\centering
\small
\setlength{\tabcolsep}{5pt}
\renewcommand{\arraystretch}{1.08}
\resizebox{\textwidth}{!}{%
\begin{tabular}{lrrrrr}
\toprule
\textbf{Region} & \textbf{Samples} & \textbf{Temperature} & \textbf{ECE} & \textbf{Multiclass Brier Score} & \textbf{Negative Log Likelihood} \\
\midrule
\multirow{6}{*}{Mainland} & 520 & 0.5 & 0.5931 & 1.1484 & 5.4307 \\
 & 520 & 0.8 & 0.5490 & 1.0633 & 3.6494 \\
 & 520 & 1.0 & 0.5250 & 1.0187 & 3.0234 \\
 & 520 & 1.2 & 0.5033 & 0.9802 & 2.6109 \\
 & 520 & 1.5 & 0.4738 & 0.9312 & 2.2117 \\
 & 520 & 2.0 & 0.4291 & 0.8671 & 1.8397 \\
\addlinespace[2pt]
\multirow{6}{*}{Singapore} & 515 & 0.5 & 0.8128 & 1.6151 & 11.0109 \\
 & 515 & 0.8 & 0.7573 & 1.5008 & 7.2449 \\
 & 515 & 1.0 & 0.7253 & 1.4368 & 5.9073 \\
 & 515 & 1.2 & 0.6966 & 1.3800 & 5.0181 \\
 & 515 & 1.5 & 0.6554 & 1.3051 & 4.1380 \\
 & 515 & 2.0 & 0.5935 & 1.2005 & 3.2960 \\
\addlinespace[2pt]
\multirow{6}{*}{Taiwan} & 536 & 0.5 & 0.6377 & 1.2748 & 10.5076 \\
 & 536 & 0.8 & 0.6278 & 1.2581 & 6.9145 \\
 & 536 & 1.0 & 0.6201 & 1.2430 & 5.5640 \\
 & 536 & 1.2 & 0.6110 & 1.2250 & 4.6519 \\
 & 536 & 1.5 & 0.5949 & 1.1934 & 3.7480 \\
 & 536 & 2.0 & 0.5621 & 1.1321 & 2.8643 \\
\addlinespace[2pt]
\multirow{6}{*}{Hong Kong} & 520 & 0.5 & 0.2820 & 0.6011 & 4.7046 \\
 & 520 & 0.8 & 0.2527 & 0.5799 & 3.2488 \\
 & 520 & 1.0 & 0.2253 & 0.5649 & 2.6808 \\
 & 520 & 1.2 & 0.2050 & 0.5508 & 2.2939 \\
 & 520 & 1.5 & 0.1774 & 0.5328 & 1.9204 \\
 & 520 & 2.0 & 0.1274 & 0.5132 & 1.5924 \\
\addlinespace[2pt]
\multirow{6}{*}{Malaysia} & 516 & 0.5 & 0.5001 & 1.0400 & 4.5044 \\
 & 516 & 0.8 & 0.4226 & 0.9274 & 3.0302 \\
 & 516 & 1.0 & 0.3800 & 0.8759 & 2.5671 \\
 & 516 & 1.2 & 0.3418 & 0.8365 & 2.2778 \\
 & 516 & 1.5 & 0.2902 & 0.7940 & 2.0173 \\
 & 516 & 2.0 & 0.2153 & 0.7525 & 1.8126 \\
\bottomrule
\end{tabular}%
}
\caption{Exact Numerical Values for Forced Choice Decoding Calibration on Qwen.}
\label{tab:decoding_calibration_qwen}
\end{table*}

\begin{table*}[t]
\centering
\small
\setlength{\tabcolsep}{5pt}
\renewcommand{\arraystretch}{1.08}
\resizebox{\textwidth}{!}{%
\begin{tabular}{lrrrrr}
\toprule
\textbf{Region} & \textbf{Samples} & \textbf{Temperature} & \textbf{ECE} & \textbf{Multiclass Brier Score} & \textbf{Negative Log Likelihood} \\
\midrule
\multirow{6}{*}{Mainland} & 520 & 0.5 & 0.5890 & 1.2787 & 3.1763 \\
 & 520 & 0.8 & 0.4157 & 1.0461 & 2.4048 \\
 & 520 & 1.0 & 0.3264 & 0.9591 & 2.2331 \\
 & 520 & 1.2 & 0.2573 & 0.9092 & 2.1612 \\
 & 520 & 1.5 & 0.1831 & 0.8737 & 2.1352 \\
 & 520 & 2.0 & 0.1091 & 0.8591 & 2.1668 \\
\addlinespace[2pt]
\multirow{6}{*}{Singapore} & 515 & 0.5 & 0.6835 & 1.4698 & 4.7808 \\
 & 515 & 0.8 & 0.5118 & 1.2183 & 3.4112 \\
 & 515 & 1.0 & 0.4240 & 1.1123 & 3.0376 \\
 & 515 & 1.2 & 0.3560 & 1.0429 & 2.8299 \\
 & 515 & 1.5 & 0.2825 & 0.9820 & 2.6671 \\
 & 515 & 2.0 & 0.2072 & 0.9372 & 2.5618 \\
\addlinespace[2pt]
\multirow{6}{*}{Taiwan} & 536 & 0.5 & 0.4377 & 1.0085 & 2.4621 \\
 & 536 & 0.8 & 0.2752 & 0.8505 & 1.9040 \\
 & 536 & 1.0 & 0.1861 & 0.7991 & 1.7989 \\
 & 536 & 1.2 & 0.1224 & 0.7752 & 1.7719 \\
 & 536 & 1.5 & 0.0444 & 0.7672 & 1.7938 \\
 & 536 & 2.0 & 0.0573 & 0.7812 & 1.8811 \\
\addlinespace[2pt]
\multirow{6}{*}{Hong Kong} & 520 & 0.5 & 0.3779 & 0.9771 & 2.7323 \\
 & 520 & 0.8 & 0.2010 & 0.8529 & 2.1772 \\
 & 520 & 1.0 & 0.1159 & 0.8211 & 2.0763 \\
 & 520 & 1.2 & 0.0637 & 0.8102 & 2.0486 \\
 & 520 & 1.5 & 0.0452 & 0.8116 & 2.0620 \\
 & 520 & 2.0 & 0.0857 & 0.8273 & 2.1258 \\
\addlinespace[2pt]
\multirow{6}{*}{Malaysia} & 516 & 0.5 & 0.4207 & 1.0121 & 2.9479 \\
 & 516 & 0.8 & 0.2750 & 0.8757 & 2.2739 \\
 & 516 & 1.0 & 0.2070 & 0.8324 & 2.1203 \\
 & 516 & 1.2 & 0.1538 & 0.8111 & 2.0547 \\
 & 516 & 1.5 & 0.0994 & 0.8015 & 2.0314 \\
 & 516 & 2.0 & 0.0578 & 0.8095 & 2.0660 \\
\bottomrule
\end{tabular}%
}
\caption{Exact Numerical Values for Forced Choice Decoding Calibration on SeaLLM.}
\label{tab:decoding_calibration_seallm}
\end{table*}

\subsection{Statistical Inference for Surface, Semantic, and Script Effects}
\label{app:surface_feature_inference}

To complement the exploratory analysis in Section~\ref{sec: Surface Feature Interference}, we 
provide formal uncertainty quantification for the surface-feature, 
semantic-domain, and script-conversion effects reported in 
Fig.~\ref{fig:Grouped Coefficient Plot}.
Rather than re-estimating these effects, our aim is to characterize their 
statistical reliability and to identify which patterns are most robust.

\paragraph{Pooled regression setup.}
We model regional Chinese understanding as item-level zero-shot 
correctness. For item $i$ and model $m$, the response variable is
$y_{i,m} = \mathbb{1}[\mathrm{parsed}_{i,m} = \mathrm{gold}_i]$.
The pooled analysis combines the three representative models used in our 
mechanistic study (Qwen3-30B, GLM-4-32B, SeaLLMs-v3-7B). 
For text and structural features, we fit a logistic regression with 
model fixed effects,
\[
P(y_{i,m}=1)
=\sigma\!\left(\beta_0 + \mathbf{z}_i^{\top}\boldsymbol{\beta} + \mu_m\right),
\]
where $\sigma(\cdot)$ is the logistic function, $\mathbf{z}_i$ denotes 
standardized surface features (Latin-script ratio, code-switch count, 
text length, and regional marker count), and $\mu_m$ captures model fixed 
effects. Each numeric feature is standardized as 
$z_{ij} = (x_{ij}-\mu_j)/\sigma_j$, so that every coefficient reflects the 
change in log-odds of correctness associated with a one-standard-deviation 
increase in the corresponding feature. 
Following the protocol of Section~\ref{sec: Surface Feature Interference}, the text and 
structural analysis covers Mainland China, Singapore, Malaysia, and 
Taiwan, yielding 6{,}261 model--item observations; Hong Kong is excluded 
because its regional-marker statistics are not directly comparable under 
the same vocabulary construction procedure.

\paragraph{Semantic-domain effects.}
For semantic domains, we report counterfactual logit shifts rather than 
dummy coefficients tied to an arbitrary reference category. 
Let the global zero-shot accuracy be $A = N^{-1}\sum_{i,m} y_{i,m}$. 
For each domain $k$, we construct a counterfactual in which all items are 
assigned to $k$ while remaining covariates are held fixed, and compute
\[
\bar{p}_k = \frac{1}{N}\sum_{i,m}\hat{P}(y_{i,m}=1 \mid d_i=k),
\]
\[
\Delta_k = \mathrm{logit}(\bar{p}_k) - \mathrm{logit}(A).
\]
Positive $\Delta_k$ indicates a domain associated with higher predicted 
correctness than the global baseline; negative $\Delta_k$ indicates 
greater predicted difficulty.

\paragraph{Uncertainty estimation.}
Standard errors, $95\%$ confidence intervals, and bootstrap $p$-values are 
obtained from $1{,}000$ bootstrap resamples. 
We additionally report Benjamini--Hochberg false discovery rate (FDR) 
adjusted $q$-values within each analysis scope. 
For script conversion, where every original item is paired with its 
script-converted counterpart, we further apply paired McNemar exact tests. 
All coefficients are interpreted as associations rather than causal effects.

\paragraph{Text and structural feature inference.}
The inference results corroborate the main-text finding that text length and localized lexical density emerge as the most robust text and structural predictors. Among the four text and structural features, text length is the only effect that remains significant after FDR correction, with a standardized logit coefficient of $+0.159$, $95\%$ CI $[0.069, 0.260]$, and adjusted $q=0.008$, indicating that items with richer contextual information are reliably associated with higher correctness. Regional marker count shows a negative association with correctness ($-0.068$, $95\%$ CI $[-0.133, -0.007]$, $q=0.064$), narrowly missing the FDR threshold but directionally consistent with the interpretation that denser region-specific lexical cues impose additional grounding demands.

The remaining features follow the directions shown in Fig.~\ref{fig:Grouped Coefficient Plot} but have uncertainty intervals overlapping zero. Latin-script ratio is negatively associated with correctness ($-0.045$, $95\%$ CI $[-0.169, 0.082]$, $q=0.626$), consistent with the interpretation that Latin-script surface forms impose additional regional-grounding demands. Code-switch count shows a near-zero aggregate coefficient ($-0.005$, $95\%$ CI $[-0.130, 0.120]$, $q=0.911$), reflecting the region-dependent direction of its effect noted in Section~\ref{sec: Surface Feature Interference}: code-mixing facilitates regional identification in Singapore and Malaysia but acts as noise in Mainland China and Taiwan, so the pooled estimate cancels out. These effects do not survive multiple-testing correction and should be read as directionally consistent rather than statistically conclusive.

\paragraph{Semantic-domain inference.}
Semantic-domain inference likewise supports the broader conclusion that 
regional Chinese difficulty is domain dependent. 
The strongest and most reliable effect appears in \emph{law, safety, and 
security}, with a counterfactual logit shift of $0.371$, bootstrap 
standard error $0.114$, $95\%$ CI $[0.153,0.585]$, and adjusted 
$q=0.030$. This suggests that institutionalized and knowledge based domains 
are associated with higher predicted correctness.

The remaining domain effects are directionally consistent with 
Section~\ref{sec: Surface Feature Interference}. 
Domains grounded in everyday local practice, including \emph{daily social life}, 
\emph{food and dining}, \emph{education}, \emph{finance and consumption}, 
\emph{transport and travel}, and \emph{work and business}, all show negative 
point estimates. In contrast, more institutional or knowledge based domains, 
including \emph{government and public services}, \emph{healthcare and 
wellbeing}, \emph{culture, history, and identity}, and \emph{technology, 
gaming, and internet}, show positive point estimates. 
Although these individual effects do not remain significant after correction, 
the consistent direction of the estimates supports the interpretation that 
models perform better on standardized domains and worse on domains that rely 
more heavily on local routines, community norms, and region specific pragmatic 
conventions.

\paragraph{Script-conversion effects.}
Script conversion is analyzed separately because its effect is 
inherently region dependent. 
For each region $r$, we compute the original and script-swapped accuracies
\[
A_{\mathrm{orig},r} = \frac{1}{N_r}\sum_i y^{\mathrm{orig}}_{i,r},
\]
\[
A_{\mathrm{swap},r} = \frac{1}{N_r}\sum_i y^{\mathrm{swap}}_{i,r},
\]
and report both the accuracy difference 
$\Delta A_r = A_{\mathrm{swap},r} - A_{\mathrm{orig},r}$ and the 
logit-scale conversion effect 
$\Delta_r = \mathrm{logit}(A_{\mathrm{swap},r}) - \mathrm{logit}(A_{\mathrm{orig},r})$.

At the aggregate level, script conversion produces only a small effect, 
consistent with the main text: simplified-to-traditional conversion 
yields a near-zero logit shift of $-0.006$, and traditional-to-simplified 
conversion a negative but non-significant shift of $-0.045$. 
This aggregate pattern, however, masks pronounced region-specific 
asymmetries.

The region-level analysis identifies Hong Kong as carrying the largest 
negative conversion effect. 
Converting Hong Kong items from traditional to simplified Chinese reduces 
accuracy from $10.19\%$ to $8.08\%$, corresponding to a logit shift of 
$-0.256$ with a $95\%$ bootstrap CI of $[-0.525,-0.022]$. 
This effect does not survive FDR correction and the paired McNemar test 
is marginal, but it provides directional evidence consistent with the 
interpretation that traditional script functions as part of a broader 
Hong Kong regional signal: when this cue is removed, the surface form 
becomes less aligned with the remaining Hong Kong-specific lexical, 
syntactic, and contact-induced written features. 
Taiwan exhibits the opposite direction, with a marginal accuracy 
increase from $61.01\%$ to $61.57\%$ after traditional-to-simplified 
conversion (logit shift $+0.024$). 
Mainland China and Malaysia show small negative effects under 
simplified-to-traditional conversion, while Singapore shows a small 
positive effect; none of these region-level intervals, apart from Hong 
Kong's, exclude zero. 
Taken together, the script-conversion results reinforce the qualitative 
conclusion of Section~\ref{sec: Surface Feature Interference}: the aggregate impact of script 
conversion is limited, but script cues interact with regional 
distributions in a non-uniform way, with the clearest degradation 
concentrated in Hong Kong and a slight positive shift in Taiwan.

Overall, the additional inference reinforces the main conclusion of 
Section~\ref{sec: Surface Feature Interference}: regional Chinese understanding failures arise 
from the joint interaction of localized lexical burden, region-dependent 
script signals, and semantic-domain dependency. 
Among the text and structural features, regional marker count is the clearest directional negative predictor, although it narrowly misses FDR correction, suggesting that dense localized lexical cues impose an additional grounding burden. The strongest statistically reliable positive domain effect is law, safety, and security, consistent with the observation that more institutionalized domains are easier for models.
Other surface, domain, and script-related effects are best interpreted 
as directionally consistent evidence that complements the exploratory 
coefficient analysis in Fig.~\ref{fig:Grouped Coefficient Plot}.

\section{Qualitative Error Case Studies}

\label{sec:error case study}

\begin{figure*}[htpb]
  \centering
  \captionsetup{skip=0pt}
  \includegraphics[width=\textwidth]{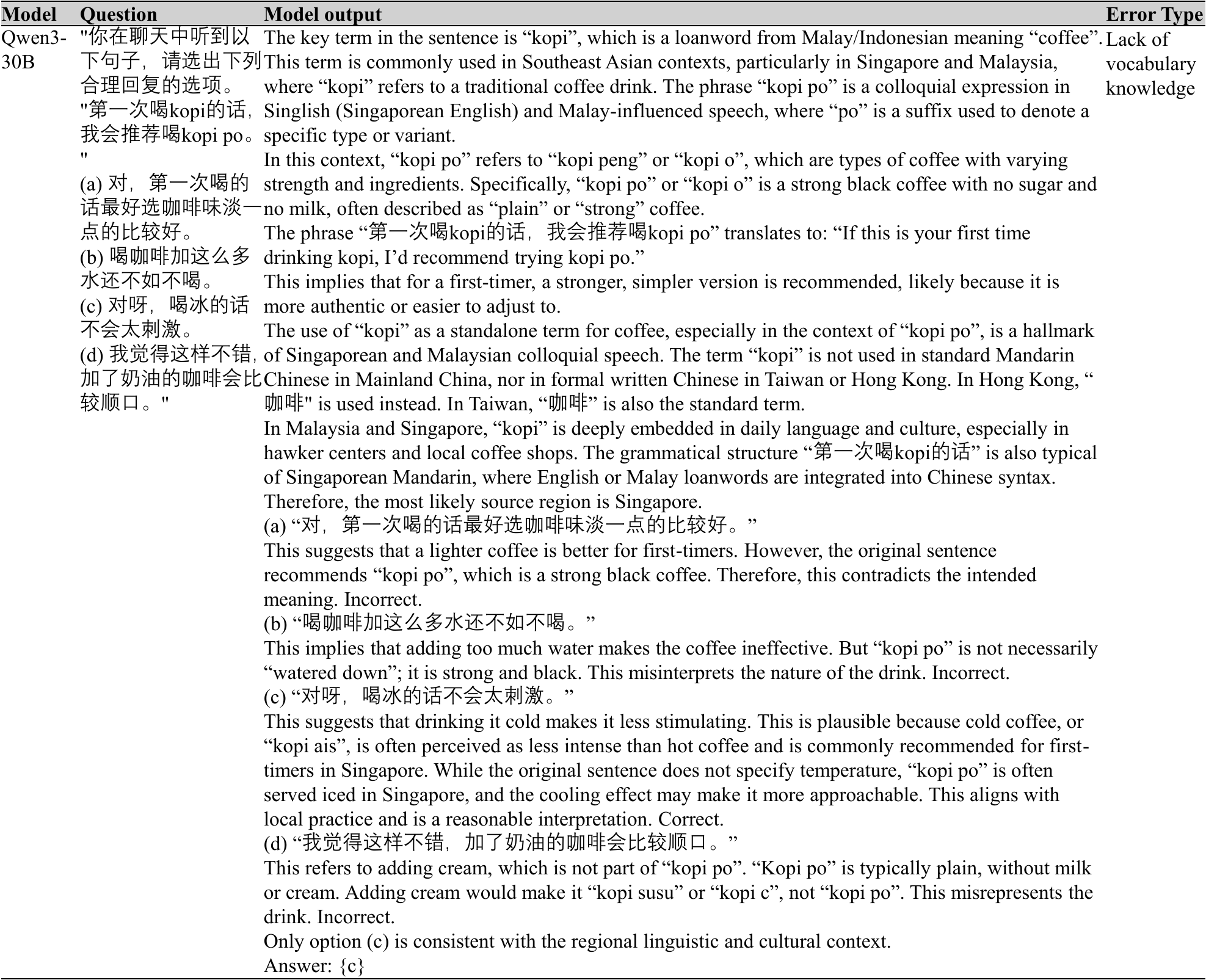}
  \caption{Qualitative examples of model reasoning failures and their corresponding error types.}
  \label{fig:knowledge gap}
\end{figure*}
\begin{figure*}[htpb]
  \centering
  \captionsetup{skip=0pt}
  \includegraphics[width=\textwidth]{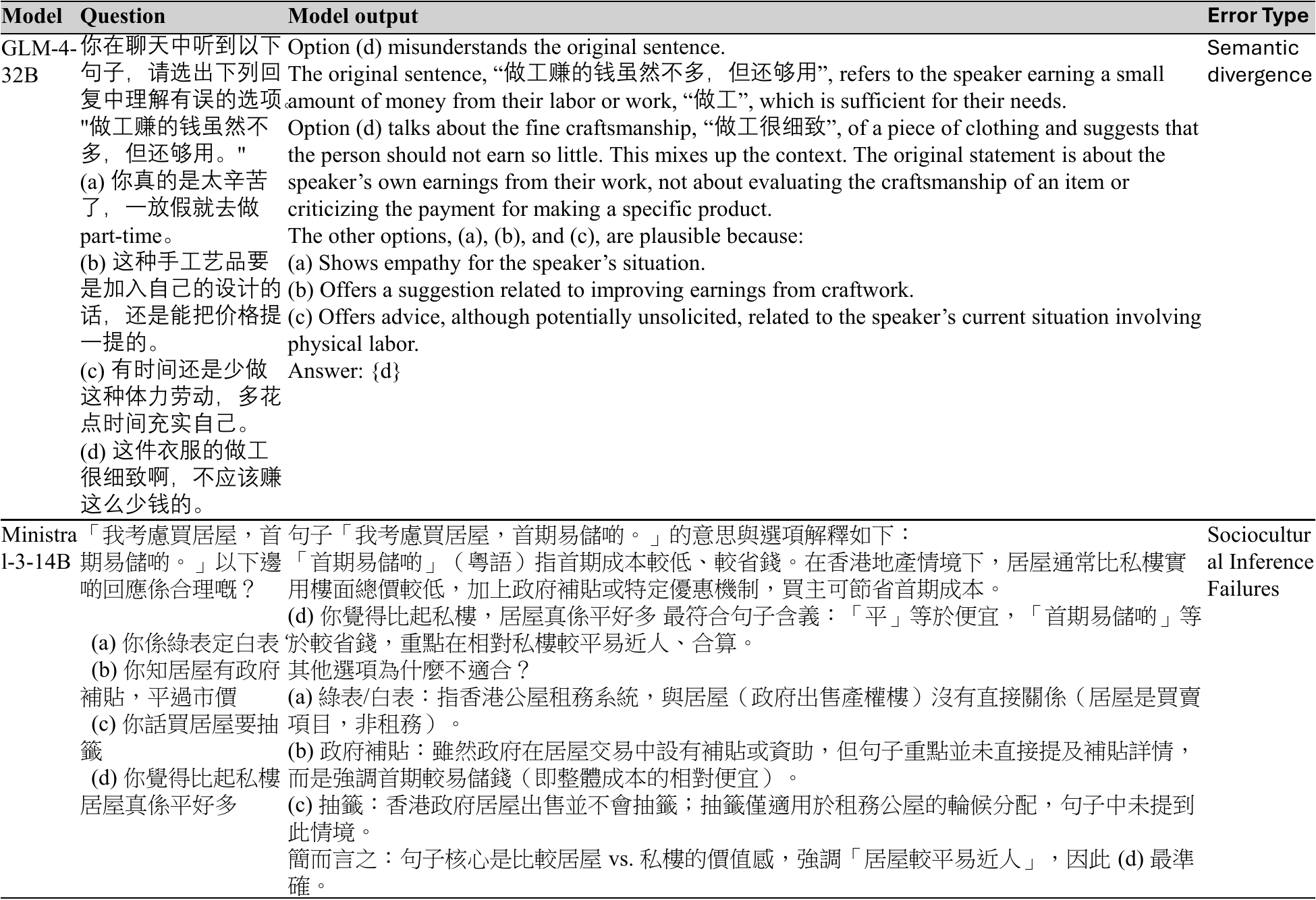}
  \caption{Qualitative examples of model reasoning failures and their corresponding error types.}
  \label{fig:semeantic sfit}
\end{figure*}
\begin{CJK}{UTF8}{gbsn}
The qualitative cases presented in Figures~\ref{fig:knowledge gap} and ~\ref{fig:semeantic sfit} further illustrate how the established error categories manifest during model reasoning. These instances demonstrate that failures in regional Chinese comprehension extend beyond simple knowledge deficits and operate instead across multiple linguistic and cognitive dimensions.

\textbf{Terminology Understanding Failures.} As illustrated by the Qwen3-30B output regarding Singaporean and Malaysian coffee terminology (see Figure~\ref{fig:knowledge gap}), direct failures in regional vocabulary grounding remain prevalent. While the model successfully identifies kopi as a localized Southeast Asian loanword for coffee, it erroneously maps ``kopi po'' to a strong black coffee and conflates it with variants such as ``kopi o'' or ``kopi peng''. This lexical hallucination propagates through the reasoning chain. The model rejects the correct response, which accurately captures the lighter flavor profile denoted by the suffix po, in favor of an irrelevant option concerning iced coffee. This case exemplifies a fundamental terminology understanding error where partial recognition of a regional marker induces a superficially plausible yet semantically invalid mapping.

\textbf{Semantic Divergence.} The GLM-4-32B output detailed in Figure~\ref{fig:knowledge gap} demonstrates the nuances of semantic divergence and highlights that errors often arise not from an absolute lack of vocabulary knowledge but from an inability to disambiguate meaning across registers. In the Singapore and Malaysia context, the term ``做工'' commonly means ``working'' or ``being employed,'' whereas in Mainland Mandarin it more typically denotes the craftsmanship or workmanship of a manufactured object. Although the model initially parses the localized meaning correctly within the premise, it fails to preserve this region-specific interpretation during candidate evaluation. Instead, it incorrectly accepts a distractor that shifts the semantic frame toward handicraft production and craftsmanship, reflecting the more standardized interpretation of the term. This vulnerability suggests that the model cannot consistently sustain region-sensitive disambiguation, allowing competing standardized senses to interfere with its reasoning process. The case therefore supports our broader observation that regional understanding errors frequently stem from unresolved semantic shifts rather than simple vocabulary gaps.

\textbf{Sociocultural Inference Failures.} Finally, the Ministral-3-14B evaluation of Hong Kong housing terminology, also shown in Figure~\ref{fig:semeantic sfit}, exposes a severe sociocultural inference failure. Although the model correctly deduces that ``居屋 '', referring to Home Ownership Scheme flats, are more affordable than private real estate, it categorically rejects valid responses referencing Green and White forms, government subsidies, and balloting procedures. By dismissing these institutionally grounded mechanisms as irrelevant or false, the model reveals that its failure is not strictly lexical. As observed in stronger models where sociocultural errors account for a larger share of remaining mistakes, the model lacks the requisite local policy knowledge to situate the utterance within the specific subsidized housing framework of Hong Kong.

These qualitative instances confirm that regional Chinese comprehension deficits operate at multiple levels. Models struggle to recover the precise semantics of localized markers, frequently conflate regionally divergent senses of otherwise familiar expressions, and routinely fail to integrate the shared social assumptions necessary for pragmatic appropriateness. Consequently, robust generalization across regional varieties demands more than superficial lexical recognition. It requires the stable grounding of localized expressions within their authentic sociolinguistic environments and institutional contexts.
\end{CJK}

\end{document}